\documentclass[final]{cvpr}

\usepackage{times}
\usepackage{epsfig}
\usepackage{graphicx}
\usepackage{amsmath}
\usepackage{amssymb}

\usepackage{latexsym}

\usepackage{xspace}
\usepackage{caption}
\usepackage{subcaption} 
\usepackage{cprotect}
\usepackage{hhline}
\usepackage{array}
\usepackage{float}
\usepackage[shortlabels]{enumitem} %

\usepackage[pagebackref=true,breaklinks=true,colorlinks,bookmarks=false]{hyperref}

\def\cvprPaperID{8131} %
\def\confYear{CVPR 2021}

\newif\ifdraft
\draftfalse

\newcommand{\coco}{COCO}
\newcommand{\cococap}{COCO Captions}
\newcommand{\oid}{Open~Images}
\newcommand{\locnarr}{LocNar}
\newcommand{\flickr}{Flickr30K}
\newcommand{\cc}{CC3M}
\newcommand{\ccp}{CC12M}
\newcommand{\nocaps}{nocaps}
\newcommand{\nocapsxd}{nocaps (XD)}
\newcommand{\sbu}{SBU Captions}
\newcommand{\visualgenome}{Visual Genome}
\newcommand{\cocorefg}{RefCOCOg}
\newcommand{\vqatwo}{VQA2}
\newcommand{\nlvrtwo}{NLVR2}
\newcommand{\ttbf}[1]{\texttt{\textbf{#1}}}

\newcommand{\cider}{{\tiny CIDEr}}
\newcommand{\spice}{{\tiny SPICE}}
\newcommand{\bleuo}{{\tiny BLEU1}}
\newcommand{\bleuf}{{\tiny BLEU4}}
\newcommand{\meteor}{{\tiny METEOR}}
\newcommand{\rouge}{{\tiny ROUGE}}

\newcommand{\ic}{{\texttt{ic}}}
\newcommand{\mass}{{\texttt{mass}}}
\newcommand{\mlm}{{\texttt{mlm}}}
\newcommand{\moc}{{\texttt{moc}}}

\newcommand{\vlm}{{\texttt{vlm}}}

\newcommand{\massrate}[1]{{\texttt{mass[#1]}}}
\newcommand{\mlmrate}[1]{{\texttt{mlm[#1]}}}

\usepackage{amssymb}
\usepackage{amsmath,amsfonts}
\usepackage{amsopn}
\usepackage{bm} %
\usepackage{multirow}
\newlength\savewidth

\newcommand{\ProbOpr}[1]{\mathbb{#1}}

\newcommand{\expect}[2]{%
\ifthenelse{\equal{#2}{}}{\ProbOpr{E}_{#1}}
{\ifthenelse{\equal{#1}{}}{\ProbOpr{E}\left[#2\right]}{\ProbOpr{E}_{#1}\left[#2\right]}}} %
\newcommand{\var}[2]{%
\ifthenelse{\equal{#2}{}}{\ProbOpr{VAR}_{#1}}
{\ifthenelse{\equal{#1}{}}{\ProbOpr{VAR}\left[#2\right]}{\ProbOpr{VAR}_{#1}\left[#2\right]}}} %

\newcommand{\eat}[1]{}

\newcommand{\mypartop}[1]{\vspace{0mm}\noindent\textbf{#1}.}
\newcommand{\mypar}[1]{\vspace{0.4em}\noindent\textbf{#1}.}

\begin{document}

\title{Conceptual 12M: Pushing Web-Scale Image-Text Pre-Training \\To Recognize Long-Tail Visual Concepts}

\author{Soravit Changpinyo, Piyush Sharma, Nan Ding, Radu Soricut\\
Google Research\\
{\tt\small schangpi,piyushsharma,dingnan,rsoricut@google.com}}

\maketitle

\begin{abstract}
The availability of large-scale image captioning and visual question answering datasets has contributed significantly to recent successes in vision-and-language pre-training. However, these datasets are often collected with overrestrictive requirements inherited from their original target tasks (e.g., image caption generation), which limit the resulting dataset scale and diversity. We take a step further in pushing the limits of vision-and-language pre-training data by relaxing the data collection pipeline used in Conceptual Captions 3M (CC3M)~\cite{cc3m} and introduce the Conceptual 12M (CC12M), a dataset with 12 million image-text pairs specifically meant to be used for vision-and-language pre-training. We perform an analysis of this dataset and benchmark its effectiveness against CC3M on multiple downstream tasks with an emphasis on long-tail visual recognition. Our results clearly illustrate the benefit of scaling up pre-training data for vision-and-language tasks, as indicated by the new state-of-the-art results on both the nocaps and Conceptual Captions benchmarks.\footnote{Our dataset is available at \url{https://github.com/google-research-datasets/conceptual-12m}.}
\end{abstract}


\begin{figure}[t]
\begin{center}
\includegraphics[width=0.49\linewidth]{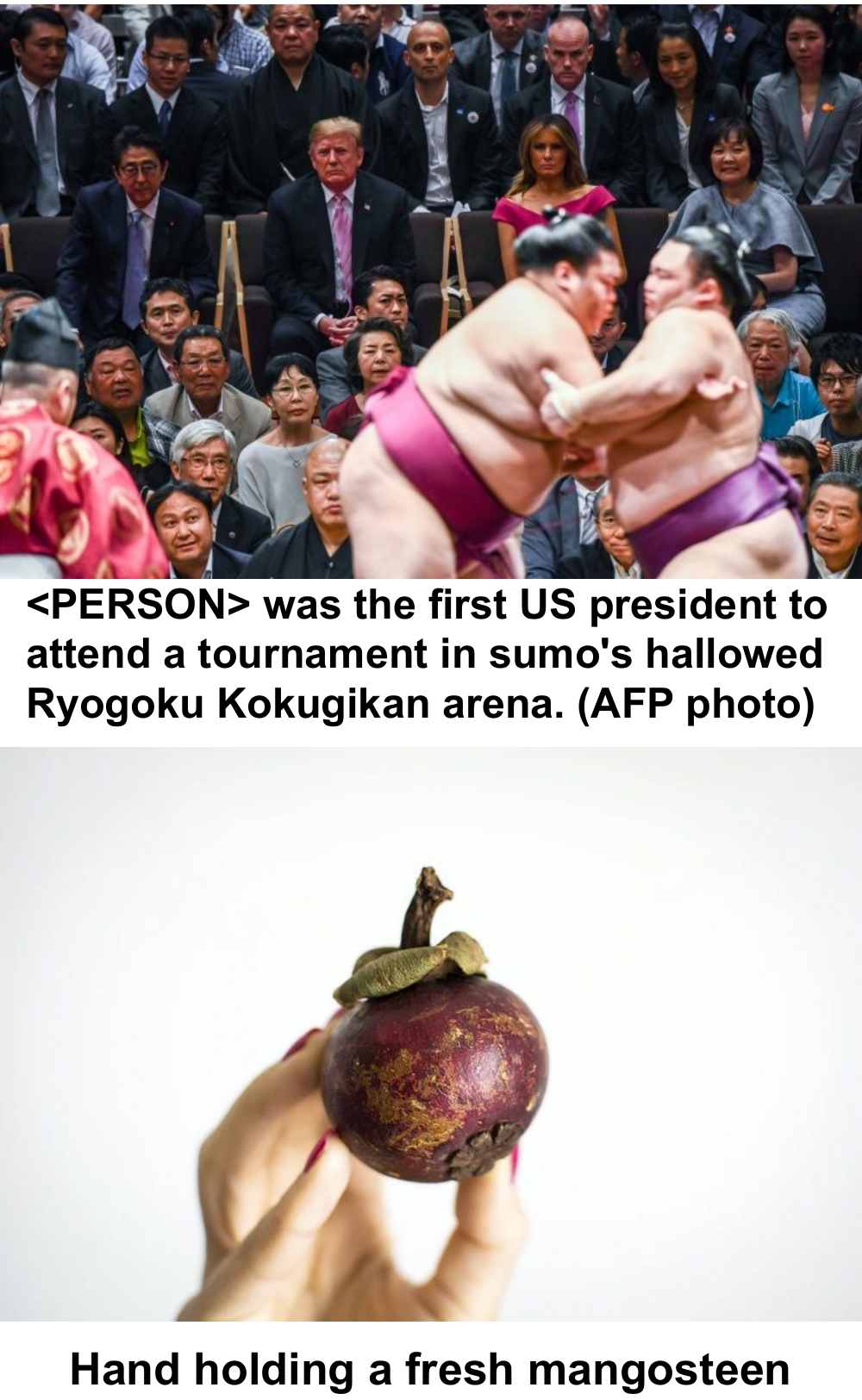}
\includegraphics[width=0.49\linewidth]{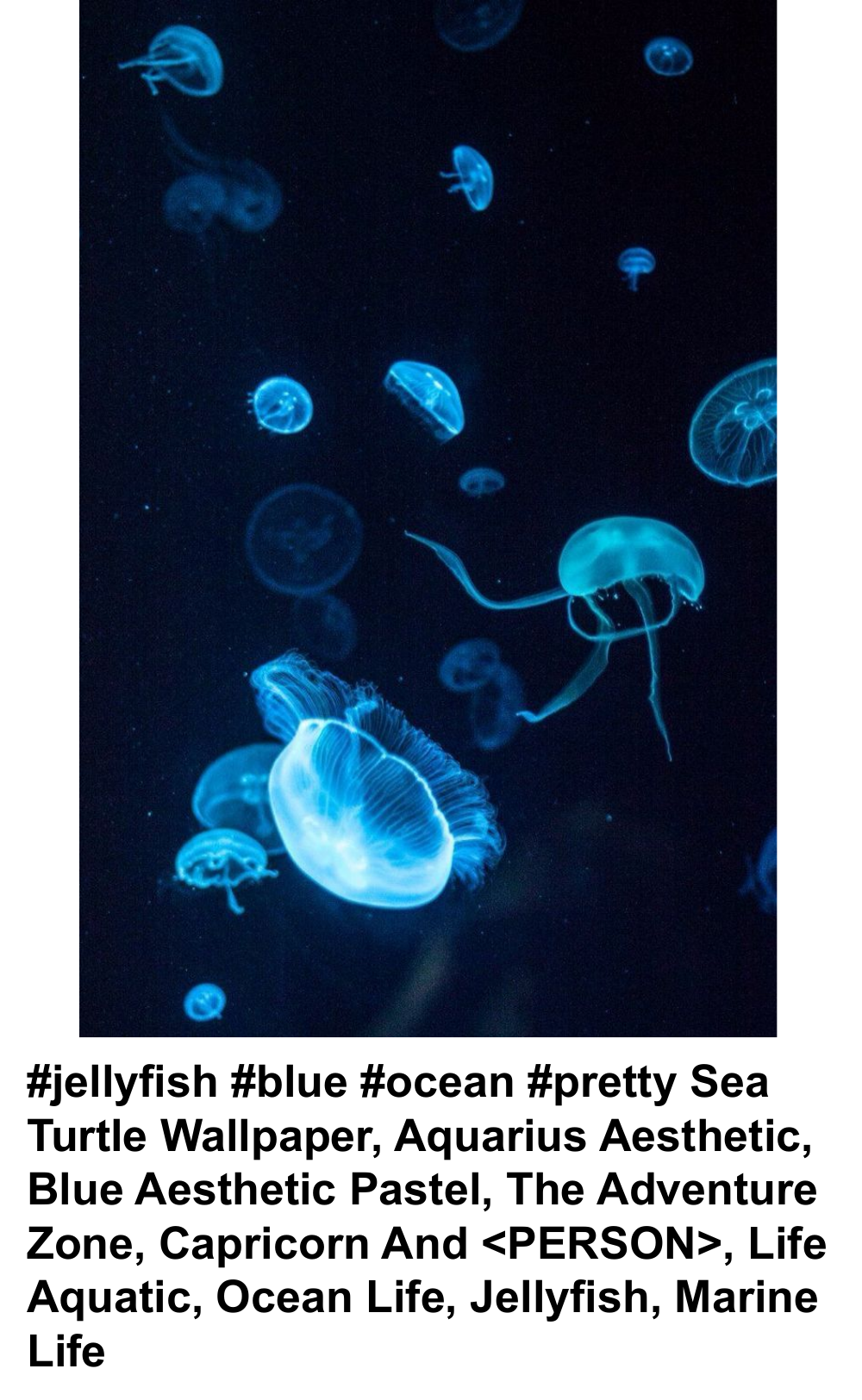}
\end{center}
 \vspace{-20pt}
 \caption{\small \textbf{\ccp{}} Even when the alt-texts do not precisely describe their corresponding Web images, they still provide rich sources for learning long-tail visual concepts such as sumo, mangosteen, and jellyfish. We scale up vision-and-language pre-training data to 12 million by relaxing overly strict filters in Conceptual Captions~\cite{cc3m}.\\}
\label{fig:cc12m}
\vspace{-30pt}
\end{figure}

\section{Introduction}
\label{sec:intro}

Transfer learning using pre-training and fine-tuning has become a prevalent paradigm in computer vision, natural language processing, and vision-and-language (V+L) research. It has been shown, for instance, that V+L pre-training leads to transferrable joint representations that benefit multiple downstream V+L tasks, including visual question answering, image and text retrieval, and referring expression comprehension~\cite{lu19vilbert,li19visualbert,chen20uniter,tan19lxmert,alberti19fusion,su20vlbert,zhou20unified,li2020unicoder,lu2012in1}.

What makes V+L pre-training successful? On one hand, this is due to advances in architectures and modeling that are mainly inspired by BERT and similar models in natural language understanding and generation~\cite{devlin19bert,liu2019roberta,yang2019xlnet,lan2020albert,goodman2019multistage,t5}.
In particular, the idea of using flexible self-attention mechanisms via high-capacity multi-layer Transformers~\cite{vaswani2017attention}, in combination with self-supervised learning objectives such as masked language modeling~\cite{devlin19bert}, has proven to be effective and widely applicable. %
On the other hand, the availability of large-scale labeled and weakly-labeled data in the V+L domain~\cite{sbucap,cococap,krishnavisualgenome,cc3m} is truly what enables such models to learn associations between the two modalities.

In \emph{either} vision \emph{or} language community, one notable trend is that scaling up training data is useful.
In contrast, datasets in V+L research remain relatively limited in terms of scale and diversity.
The capability of JFT-300M~\cite{jft300m} and Instagram~\cite{instagram} over orders-of-magnitude smaller ImageNet~\cite{imagenet15} has been put to test on multiple downstream image classification and object detection tasks.
In NLP, the size of pre-training data sources for training deep language models rose from the 20GB BooksCorpus~\cite{zhu15aligning}+English Wikipedia in BERT\cite{devlin19bert}, to the 570GB dataset in GPT-3~\cite{gpt3} and the 745GB C4 dataset in T5~\cite{t5}. 

In contrast, V+L datasets are limited in two ways. First, the \emph{effective} sizes of popular V+L datasets are low. The number of images in these datasets range from fewer than a few hundred thousands~\cite{flickr,cococap,krishna2017visual,vizwizcap} to several millions~\cite{cc3m}, with lower text quality as the scale increases.
Second, many of the small-sized datasets share the same, limited visual domain; COCO-Captions~\cite{cococap}, Visual Genome~\cite{krishna2017visual}, and VQA2~\cite{goyal2017making} are (mostly) based on several hundreds thousand of COCO images~\cite{coco}.
The lack in scale and diversity of visual concepts (with respect to vision/language-only counterparts) makes it hard for V+L models to perform adequately in the wild.

One major reason for these gaps is the difficulty in collecting such datasets. Unlike in image classification, ``text" in V+L datasets is longer and less likely to be agreed upon, making the annotation process more costly and time-consuming. One approach to remedy this is to make use of large amounts of the alt-texts accompanying images on the Web. For instance, Sharma et al. introduced Conceptual Captions (\cc{})~\cite{cc3m}, a dataset of 3.3M $\langle$image, caption$\rangle$ pairs that result from a filtering and postprocessing pipeline of those alt-texts. Despite being automatically collected, \cc{} is shown to be effective in both image captioning in the wild~\cite{cc3m,changpinyo2019decoupled} and V+L pre-training~\cite{lu19vilbert,li19visualbert,chen20uniter,tan19lxmert,alberti19fusion,su20vlbert,zhou20unified,li2020unicoder,lu2012in1}. In other words, it provides a promising start for large-scale V+L annotations.

In this paper, we explore pushing the limits of V+L data using this approach.
Our key insight is that specific downstream V+L tasks (e.g., VQA, image captioning) can be overly restrictive if the goal is to collect large-scale V+L annotations.
For instance, \cc{} was collected to favor high-precision texts that are fit for the downstream task of image captioning.
Yet, we have witnessed this dataset being increasingly adopted for V+L
pre-training~\cite{lu19vilbert,chen20uniter,alberti19fusion,su20vlbert,zhou20unified,li2020unicoder,lu2012in1},
arguably beyond its original purpose.

We hypothesize that the V+L field could benefit from such an insight, and therefore we introduce Conceptual 12M (\ccp{}), a high(er)-recall V+L dataset for the purpose of V+L pre-training.
By relaxing multiple image and text filters used in \cc{}, we obtain a less precise but 4x larger V+L set of $\langle$image, text$\rangle$ pairs. We perform an analysis of this dataset and show that it covers a wider range of visual concepts.

We test our hypothesis by benchmarking the effectiveness of \ccp{} as a pre-training data source on several V+L tasks, in comparison to \cc{}. We explore two main pre-training strategies (and more in the Supplementary material): one for vision-to-language generation and the other for vision-and-language matching. Our experiments indicate that scaling up pre-training V+L has a dramatic positive effect on image captioning, novel object captioning, and (zero-shot) image retrieval.

In summary, our main contributions are:
\begin{enumerate}[(a),nosep,leftmargin =* , widest* = 8]
  \item A public larger-scale V+L pre-training dataset that covers a much wider range of concepts than existing ones. 
  \item Evaluation on downstream vision-to-language generation and vision-and-language matching with an emphasis on long-tail recognition that consistently shows the superiority of this dataset over \cc{}.
  \item State-of-the-art results on the \nocaps{} (novel object captioning) and Conceptual Captions benchmarks.
\end{enumerate}


\section{Vision-and-Language Pre-Training Data}
\label{sec:data}

We first review the data collection pipeline for the Conceptual Captions 3M (\cc{}) outlined in Sect.~3 of \cite{cc3m}, which we followed closely. We then describe a series of relaxation and simplification to the pipeline that results in \ccp{}, a much larger set of image-text pairs. Finally, we perform an analysis of \ccp{} in comparison with \cc{} and other existing V+L datasets.

\subsection{Conceptual Captions 3M: Pipeline for extracting and cleaning Image Alt-Text from the Web}

The Conceptual Captions dataset consists of about 3.3M Web images and their corresponding cleaned, hypernymized Alt-texts~\cite{cc3m}. 
This approach leverages a promising source of (weak) supervision for learning correspondance between visual and linguistic concepts:
once the pipeline is established, the data collection requires no additional human intervention. 
It consists of the following 4 steps: 
(i) image-based filtering based on size, aspect ratio, encoding format and offensive content, 
(ii) text-based filtering based on language, captialization, token frequency, pre-defined unwanted phrases, as well as part-of-speech (POS), sentiment/polarity, and adult content detection (using Google Cloud Natural Language APIs), 
(iii) image-text--based filtering based on the number of image tags (as predicted by Google Cloud Vision APIs) that overlap with the existing text, 
(iv) text transformations, most notably hypernymization of named entities, including proper names of persons, organizations and locations (e.g., both “Harrison Ford” and “Calista Flockhart” are replaced by ``actor"), deletion of time-related spans, and digit replacement (using \# as a digit abstraction).

The large scale nature and the high degree of textual and visual diversity make this dataset particularly suited to V+L pre-training~\cite{lu19vilbert,chen20uniter,su20vlbert,zhou20unified,li2020unicoder,lu2012in1}.

\subsection{\ccp{}: Relaxing filters for higher recall}
\label{sec:data_relax}

Conceptual Captions has been created to work out-of-the-box for training image captioning models, and thus it involves substantial image, text, and image-text filtering and processing to obtain clean, high-precision captions.
As a result, this approach comes at the cost of low recall (many potentially useful $\langle$image, Alt-text$\rangle$ pairs are discarded).
However, this trade-off may not be optimal if the dataset is to be used primarily for V+L pre-training.
Motivated by this, we follow a similar procedure as the one described in \cite{cc3m} but relax some of its filters, and construct the dataset called Conceptual 12M (\ccp{}), as detailed below.

\mypar{Filtering}
As described above, the construction of \cc{} used three main filtering types~\cite{cc3m}: image-based, text-based, and image-text--based. 
To arrive at \ccp{}, we keep the image-text filtering intact, and relax the unimodal filters only.
First, for image-based filtering, we set the maximum ratio of larger to smaller dimension to 2.5 instead of 2.
We still keep only JPEG images with size greater than 400 pixels, and still exclude images that trigger pornography detectors.
Second, in text-based filtering, we allow text between 3 and 256 words in the alt-text.
We still discard candidates with no noun or no determiner, but permit ones without prepositions.
We discard the heuristics regarding high unique-word ratio covering various POS tags and word capitalization.
We set the maximum fraction of word repetition allowed to 0.2.
Given a larger pool of text due to the above relaxations, the threshold for counting a word type as rare is increased from 5 to 20.

\mypar{Text transformation}
The main motivation for \cc{} to perform text transformation is that a majority of candidate captions contain ultrafine-grained entities such as proper names (people, venues, locations, etc.), making it extremely difficult to learn as part of the image captioning task. 
In contrast, we are not restricted by the end task of image caption generation. Our intuition is that relatively more difficult pre-training data would lead to better transferability. 
We thus do not perform hypernimization or digit substitution as in \cite{cc3m}. 
The only exception to the ``keep alt-texts as raw as possible" rule is performing person-name substitutions, which we identify as necessary to protect the privacy of the individuals in these images. 
For this step, we use the Google Cloud Natural Language APIs to detect all named entities of type Person, and substitute them by a special token $\langle$PERSON$\rangle$.
Around 25\% of all the alt-texts in \ccp{} are transformed in this fashion.

\begin{figure}[t]
\begin{center}
\includegraphics[width=0.7\linewidth]{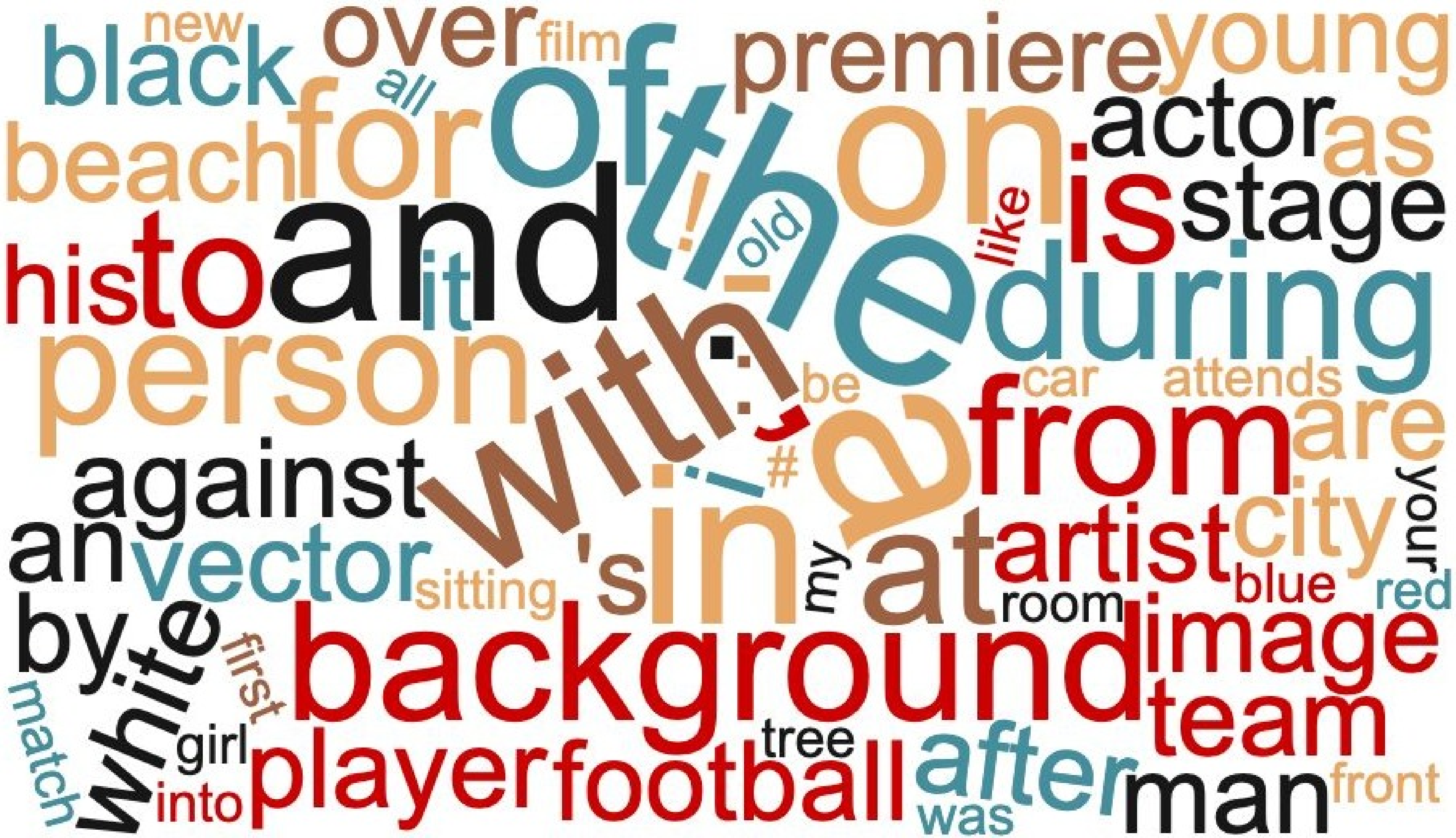}
\par \vspace{3pt}
\includegraphics[width=0.7\linewidth]{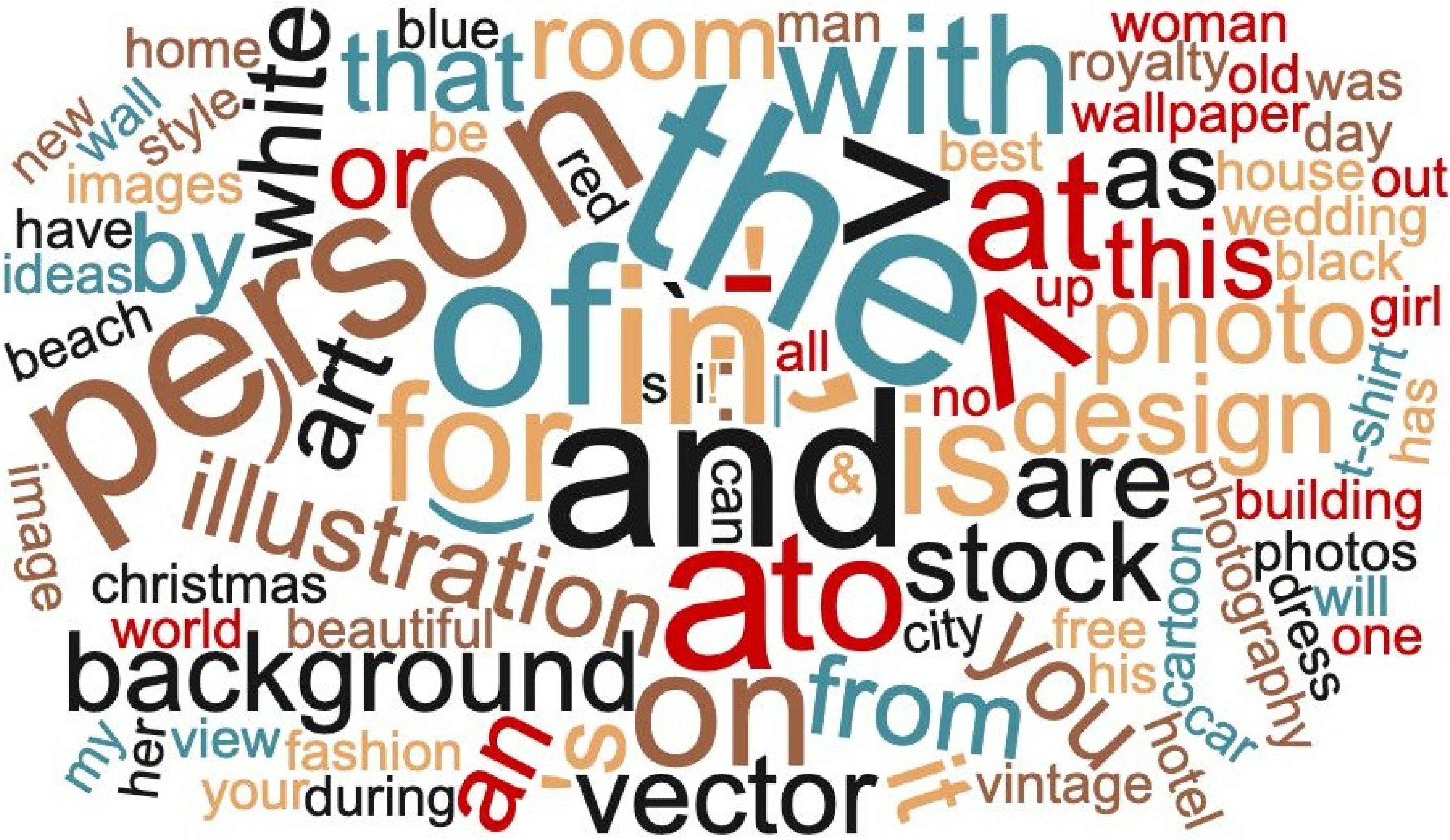}
\end{center}
 \vspace{-20pt}
 \caption{\small Word clouds of top 100 tokens in \cc{} (the top cloud) and in \ccp{} (the bottom cloud).\\}
\label{fig:wordclouds}
\vspace{-30pt}
\end{figure}

\begin{table}[t]
\small
\begin{center}
\begin{tabular}{c|c|c|c}
Dataset & \# examples & token/type & caption length \\ \hline
\cc{} train & 3,318,333 & 804.8 & 10.3 $\pm$ 4.5 \\
\ccp{} & 12,423,374 & 370.0 & 20.2 $\pm$ 16.3  \\ \hline
\end{tabular}
\vspace{-5pt}
\caption {\small Basic statistics of \ccp{} vs. \cc{}}
\vspace{-30pt}
\label{tab:datastats}
\end{center}
\end{table}

\subsection{Characteristics of \ccp{}}
\label{sec:data_char}

We provide an analysis of \ccp{} along multiple dimensions, focusing on comparing it to the most relevant \cc{}. Additional analyses are in the supplementary material. 

\mypar{Basic statistics}
As seen in Table~\ref{tab:datastats},
\ccp{} consists of 12.4M image-text pairs\footnote{Extracted as of May 2020.}, about 4x larger than \cc{}.
It has a much lower token (word count) to type (vocab size) ratio, indicating a longer-tail distribution and a higher diversity degree of the concepts captured.
Lastly, the average caption length of \ccp{} is much longer.
This is overall achieved by our relaxation of the filters, especially the text one.

\mypar{Quality}
We compute a rough estimate of precision on 100 examples by asking two annotators to rate how well the given alt-text fits the image on a 1--5 scale: 1 (no fit), 2 (barely fit), 3 (somewhat), 4 (good fit, but disfluent language), 5 (perfect).
We define precision as the fraction of captions with a score 4 or above. We see a drop in precision, 76.6\% vs. 90.3\% as reported for \cc{} (Table~2 in \cite{cc3m}).
This analysis points to the precision/recall tradeoff in transitioning from \cc{} to \ccp{}.
Fig.~\ref{fig:cc12m} illustrates such a tradeoff: the ``jellyfish'' example would have been filtered out from \cc{} (due to a high percentage of nouns and a lack of proprositions), but it is included in \ccp{}.

\mypar{Visual concept distribution}
We use the caption text tokens to represent the visual concepts. The long tail of visual concepts that emerge in \ccp{} spans many categories, and can be attributed to 
(1) a dramatic increase in scale, and 
(2) the absence of fine-grained entity hypernymization.
We list some of them here to illustrate this point, in the format of ``$\langle$word$\rangle$ $\langle$frequency in \cc{}$\rangle$ $\xrightarrow{}$ $\langle$frequency in \ccp{}$\rangle$":
luffy 0 $\xrightarrow{}$ 152,
mangosteen 0 $\xrightarrow{}$ 212,
zanzibar 0 $\xrightarrow{}$ 1138,
sumo 1 $\xrightarrow{}$ 661,
pokemon 1 $\xrightarrow{}$ 8615,
chevrolet 1 $\xrightarrow{}$ 12181,
mehndi 3 $\xrightarrow{}$ 9218,
pooh 4 $\xrightarrow{}$ 7286,
cyberpunk 5 $\xrightarrow{}$ 5247,
keto 6 $\xrightarrow{}$ 6046,
hound 9 $\xrightarrow{}$ 3392,
quiche 50 $\xrightarrow{}$ 1109,
durian 61 $\xrightarrow{}$ 552,
jellyfish 456 $\xrightarrow{}$ 2901.

We also visualize the head of the distribution in Fig.~\ref{fig:wordclouds}. We observe that ``person" becomes much more frequent due to person substitution with the token ``$\langle$PERSON$\rangle$". Moreover, there are fewer ``actor", ``artist", ``(football) player", as a result of removing hypernymization.

Finally, we inspect tokens that are unseen in \cc{}. We observe that these tokens may occur very frequently in \ccp{} if they are fine-grained instances such as locations (``france," ``africa," ``dc," ``toronto") or digits (``2019", ``10", ``2018", ``2020").
This is due to the removal of hypernymization and the dropping of time-related span deletion.

\mypar{Biases}
We study the context in which several sensitive terms related to gender, age, race, ethnicity appear such as ``black"/``white"/``asian"/``african"/``indian", ``man"/"woman", ``young"/``old", etc. We observe no large biases in the distribution of these terms, either in terms of co-occurrence between sensitive term pairs or with other tokens.
Furthermore, we check the distribution of web domains and, similar to visual concepts, we find this to be diverse and long-tail: $>$100K with $>$40K contributing $>$10 samples. We take our preliminary study as a positive indication of no severe biases stemming from particular domains or communities. Finally, we provide a Broader Impact statement in the supplementary material. 


\begin{figure}[t]
\begin{center}
\includegraphics[width=0.92\linewidth]{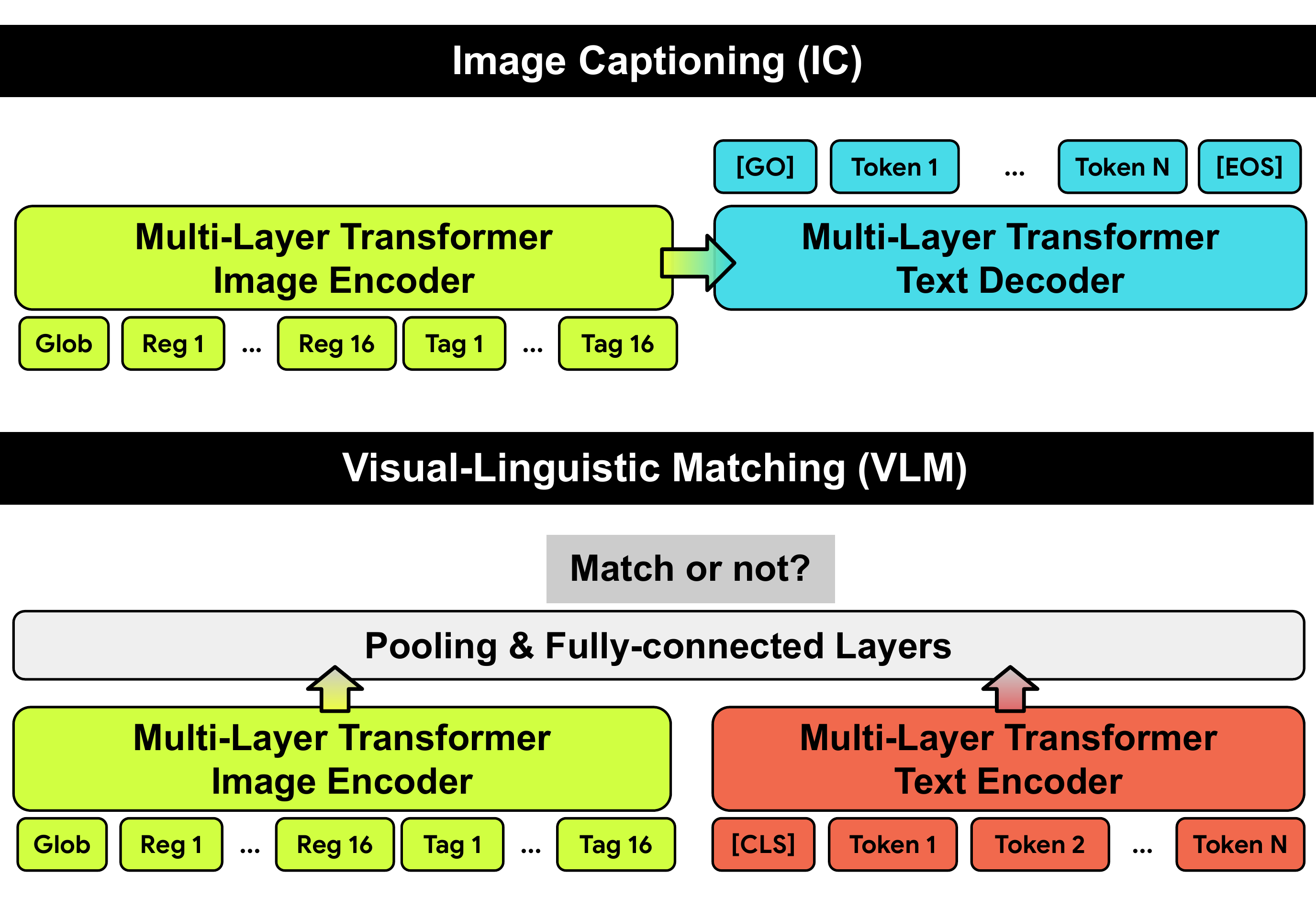}
\end{center}
 \vspace{-15pt}
 \caption{\small \textbf{Main Pre-Training Tasks:} image captioning (vision-to-language generation) and visual-linguistic matching (vision-and-language understanding).}
\label{fig:pretraintasks}
\vspace{-15pt}
\end{figure}

\begin{table}[t]
\small
\begin{center}
\begin{tabular}{c|c|c}
Downstream & \multicolumn{2}{c}{Downstream datasets} \\ \cline{2-3}
task       & Train & Eval \\ \hline
Novel object captioning & \cococap{} & \nocaps{} \\
Novel object captioning & \locnarr{} COCO & \locnarr{} OID \\
Image captioning & \multicolumn{2}{c}{\cc{}} \\ \hline
Zero-shot IR & None & \flickr{} \\ 
IR & \multicolumn{2}{c}{\flickr{}} \\
IR & \multicolumn{2}{c}{\locnarr{} \flickr{}} \\ \hline
\end{tabular}
\vspace{-5pt}
\caption {\small Generation (top) and matching (bottom) tasks and datasets considered in this paper. IR = Image Retrieval.}
\vspace{-30pt}
\label{tab:tasks}
\end{center}
\end{table}

\section{Evaluating Vision-and-Language Pre-Training Data}
\label{sec:exp_setup}

The previous section describes \cc{} and our \ccp{}. In this section, we evaluate both datasets on their ability to benefit V+L downstream tasks, measuring the impact from visual grounding produced under the two settings. For the sake of comparison, we do not include the images that appear in \cc{} in \ccp{} in our experiments.

We focus on the two most fundamental V+L tasks: vision-to-language \textbf{generation} and vision-and-language \textbf{matching}. In both cases, our emphasis is on
(i) the simplest setting in which the learning objectives during pre-training and downstream tasks match, and
(ii) long-tail recognition and out-of-distribution generalization, as we believe this is where pre-training has the most impact. Fig.~\ref{fig:pretraintasks} and Table~\ref{tab:tasks} summarize our experimental setup, in terms of the downstream tasks and the fine-tuning and evaluation datasets.

\subsection{Vision-to-Language Generation}
\label{sec:gen}

\subsubsection{Pre-Training Tasks}

We use \textbf{image captioning (\ic)} as the pre-training task.
The task is to predict the target caption given image features.
To train the model parameters, we use the standard cross entropy loss given the groundtruth caption.

Note that there exist vision-to-language generation pre-training strategies that are different from ours. 
For instance, Zhou et al.~\cite{zhou20unified} adapt BERT~\cite{devlin19bert} to generate text. As masked language modeling is used for pre-training, there is no decoder and, at inference time, text is generated using the encoder network one token at a time, appending the mask token to the image and the text generated so far. Thus, this approach is inefficient as the number of passes over the input image is linear in the desired caption length. It is also unclear how to incorporate advanced decoding schemes such as beam search, top-k sampling, or nucleus sampling (see, e.g., \cite{holtzman20thecurious}) with such an approach.
Finally, our experiments (see Supplementary material) suggest that the \ic{} pre-training task is superior to its masked variants and justify using the simple \ic{} learning objective.

\subsubsection{Downstream Tasks} \label{sec:gen_downstream}
Our downstream tasks are selected to measure progress toward solving image captioning in the wild.
They also stand to benefit from visual grounding, especially since pre-training, by definition, is expected to cover a wider range of (long-tail) visual concepts than fine-tuning datasets.

\textbf{\nocaps~\cite{nocaps}} is a recent object-captioning-at-scale benchmark consisting of 4,500 validation and 10,600 test images with 10 hidden reference captions.
Unlike in the standard image captioning setting, \nocaps{}'s distributions of images during training (\cococap{}) and evaluation (\oid{}) are different: the \oid{} dataset~\cite{oidv4,openimages} covers one order of magnitude more objects (600 classes) than \coco~\cite{coco} (80 classes).
This discrepancy defines the challenge: solutions must be able to learn to describe novel concepts from sources external to the COCO training set, such as text corpora, knowledge bases, or object detection datasets.
In the Supplementary material, we provide details on the \nocaps leaderboard. In addition, besides \cc{} and \ccp{}, we also explore using the Open Images Localized Narratives dataset (\locnarr{}) \cite{locnarr}, as an alternative ``in-domain" (from a visual standpoint) pre-training data source.

\textbf{Localized Narratives (\locnarr{})~\cite{locnarr}} is a collection of datasets with images that are paired with captions obtained by converting speech to text via ASR and manual post-processing it\footnote{This dataset also contains mouse traces synchronized with the text, but we do not use the traces here.}. Inspired by the setting in \nocaps{}, we use the \coco \cite{coco} portion (train split of around 130K images) for training/fine-tuning, and \oid{}~\cite{oidv4} portion of evaluation (val split of around 40K images).
Note that the LocNar captions are much longer than standard captioning datasets (41.8 words/caption), setting it apart from \nocaps{}. 

\textbf{Conceptual Captions 3M~\cite{cc3m}} is our main reference for V+L pre-training data source.
At the same time, the image captioning task on this dataset itself is a valuable benchmark for vision-to-language generation in the wild. Thus, we adopt it as a downstream task for \ccp{}.
This means that, in the case of \cc{}, from-scratch and pre-training settings collapse.

\mypar{Evaluation metrics}
To measure the performance on image caption generation, we consider the standard metrics BLEU-1,4~\cite{bleu}, ROUGE-L~\cite{rouge}, METEOR~\cite{meteor}, \mbox{CIDEr-D}~\cite{cider}, and SPICE~\cite{spice}.

\subsection{Vision-and-Language Matching}

\subsubsection{Pre-training Tasks}
In \textbf{visual-linguistic matching (\vlm)}, the task takes as input both image and text features and predicts whether the input image and text are matched.
To train the model's parameters, we use a contrastive softmax loss, for which the original image-text pairs are used as positive examples, while all other image-text pairs in the mini-batch are used as negative examples~\cite{lu19vilbert,tan19lxmert}.

\subsubsection{Downstream Tasks}
The task of \textbf{caption-based image retrieval (IR)} is to identify a relevant image from a pool given a caption describing its content.
The \flickr{} dataset \cite{flickr30k} consists of 31,000 images from Flickr, each associated with five captions.
Following existing work \cite{lee18stacked,lu19vilbert}, we use 1,000 images for validation, 1,000 images for testing, and use the rest of image-text pairs for model training.

We further consider \textbf{zero-shot caption-based image retrieval}~\cite{lu19vilbert} on the \flickr{} dataset.
The term ``zero-shot" refers to the setting in which we discard training data and apply pre-trained models ``as-is'', i.e., without fine-tuning on the target task.

Finally, we further evaluate our retrieval system on the Localized Narratives dataset \cite{locnarr} (see Sect.~\ref{sec:gen_downstream}). We use the \locnarr{} \flickr{} portion (train split of 30,546 images, and test split of 1000 images) for training and evaluation.

\mypar{Evaluation metrics}
To measure the performance on image retrieval, we consider the standard metrics Recall@1 (R1), Recall@5 (R5), and Recall@10 (R10).

\begin{table*}[t]
\small
\begin{center}
\begin{tabular}{r|c|cc|cc|cc|cccccc|}
\multicolumn{1}{c|}{Pretraining} & Train or & \multicolumn{12}{c}{\ttbf{nocaps val}} \\ \cline{3-14}
\multicolumn{1}{c|}{data} & finetune on & \multicolumn{2}{c|}{in-domain} & \multicolumn{2}{c|}{near-domain} &  \multicolumn{2}{c|}{out-of-domain} & \multicolumn{6}{c|}{overall}\\ \cline{3-14}
 & coco cap? & \cider & \spice & \cider & \spice & \cider & \spice & \bleuo & \bleuf & \meteor & \rouge & \cider & \spice \\ \hline
None & \checkmark & 72.8 & 11.1 & 57.1 & 10.2 & 34.1 & 8.3 & 69.8 & 14.5 & 21.9 & 47.9 & 54.7 & 10.0 \\ \hline
\cc{} &  & 29.2 & 7.4 & 27.5 & 6.9 & 37.3 & 7.4 & 36.0 & 2.8 & 12.6 & 29.1 & 29.7 & 7.1 \\
\ccp{} &  & 20.7 & 6.9 & 24.1 & 6.9 & 41.6 & 8.0 & 31.8 & 2.9 & 12.1 & 26.8 & 27.1 & 7.2 \\ \hline
\cc{} & \checkmark & 81.8 & 11.6 & 73.7 & 11.1 & 65.3 & 10.1 & 74.6 & 19.1 & 24.1 & 51.5 & 73.2 & 11.0 \\
\ccp{} & \checkmark & \underline{88.3} & \underline{12.3} & \underline{86.0} & \underline{11.8} & \underline{91.3} & \underline{11.2} & \underline{78.5} & \underline{23.4} & \underline{25.9} & \underline{54.5} & \underline{87.4} & \underline{11.8} \\
\hline
\cc{}+\ccp{} & \checkmark & \textbf{92.6} & \textbf{12.5} & \textbf{88.3} & \textbf{12.1} & \textbf{94.5} & \textbf{11.9} & \textbf{79.2} & \textbf{24.4} & \textbf{26.1} & \textbf{55.1} & \textbf{90.2} & \textbf{12.1} \\ \hline
\end{tabular}
\vspace{-7pt}
\caption {\small Automatic metric scores on the \nocaps{} val set: performance of from-scratch (Row 1), pre-trained (Rows 2-3), and fine-tuned (Rows 4-5) models. \ccp{} outperforms \cc{} by a large margin after fine-tuning (Row 4 vs. 5). Together, they achieve a new best, surpassing 90 CIDEr points on nocaps val. Bold indicates best-to-date, underline indicates second-best.}
\vspace{-20pt}
\label{tab:nocaps}
\end{center}
\end{table*}

\begin{figure*}[t]
\begin{center}
\includegraphics[width=0.33\linewidth]{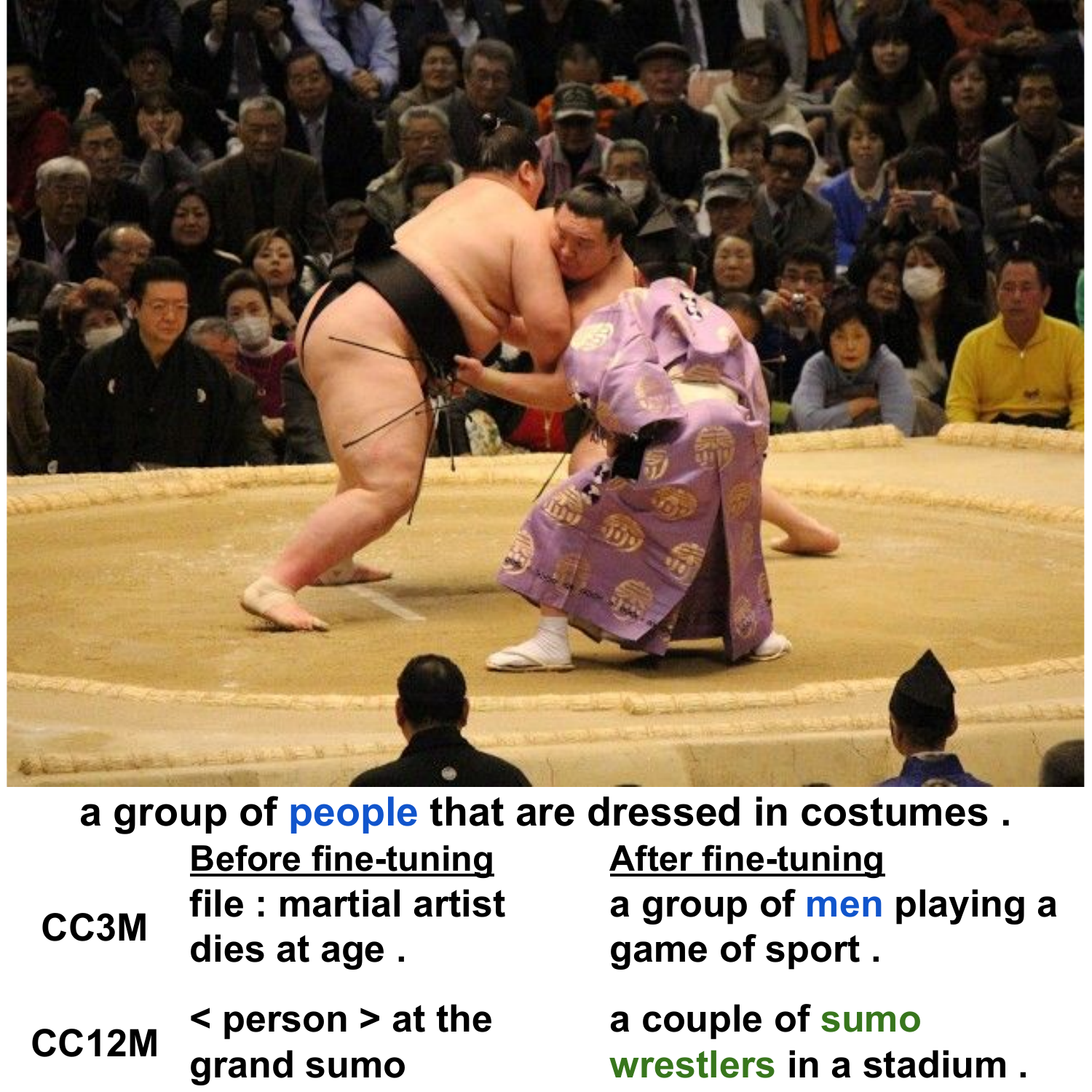}
\includegraphics[width=0.33\linewidth]{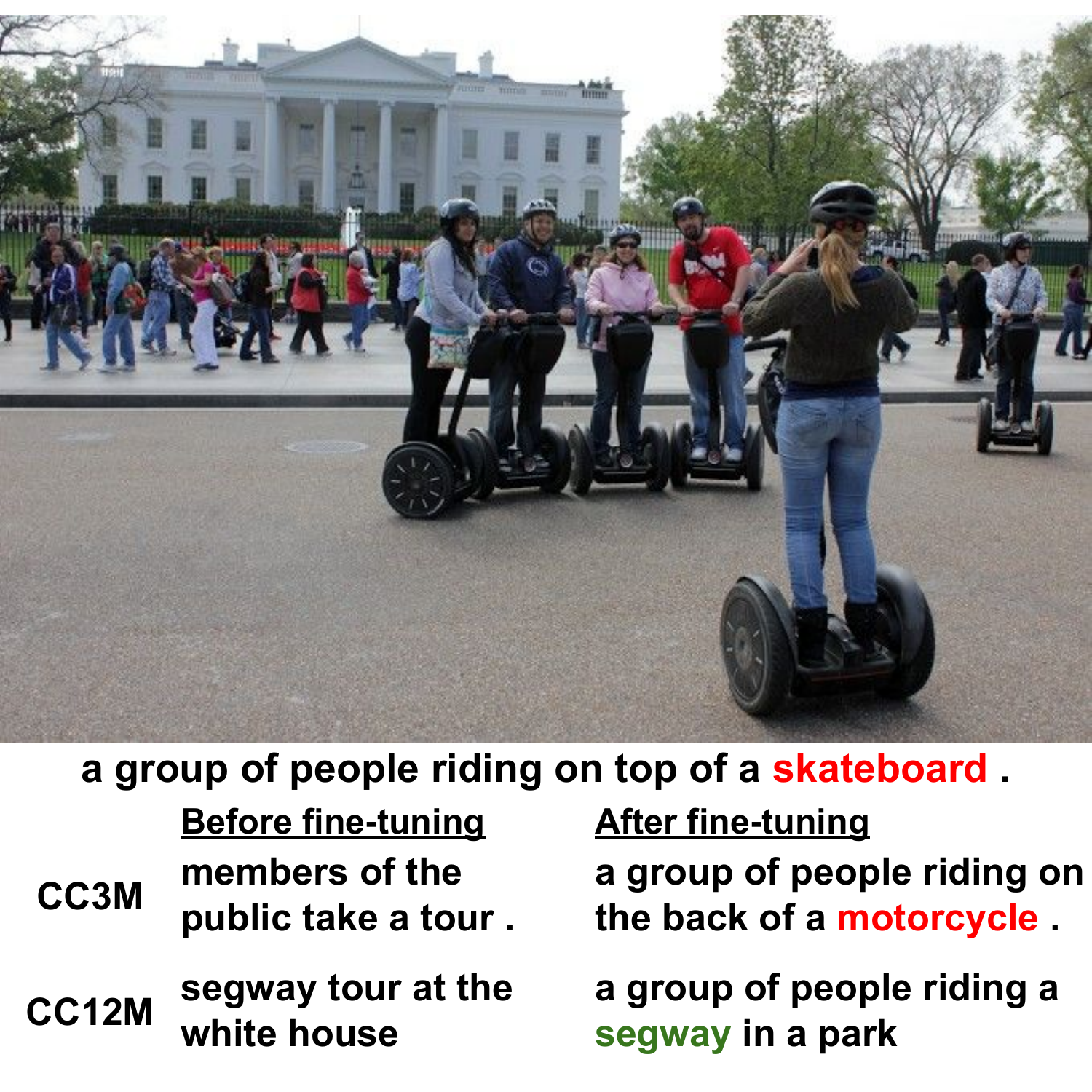}
\includegraphics[width=0.33\linewidth]{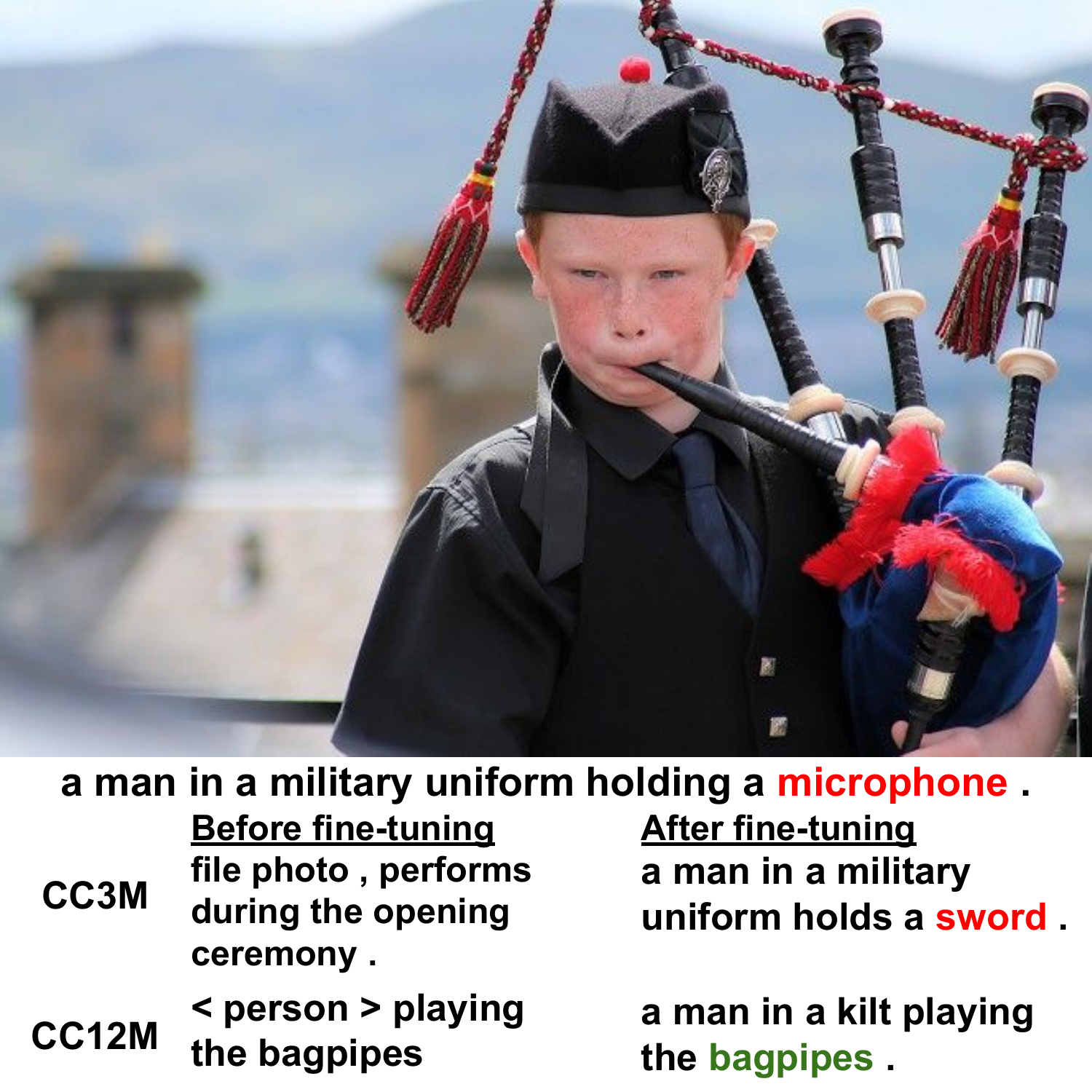}
\end{center}
 \vspace{-12pt}
 \caption{\small \textbf{Qualitative results on nocaps}. Each example comes with a caption predicted by the model that is trained on \cococap{} without pre-training (very top, right under the image), as well as captions predicted by models pre-trained on \cc{} (middle) and \ccp{} (bottom), where the left/right column indicates if the model is fine-tuned on \cococap{}.}
\label{fig:nocaps_qual}
\vspace{-15pt}
\end{figure*}

\subsection{Implementation Details}

\mypartop{Representing images and texts}
We use Graph-RISE~\cite{juan19graphrise,juan20ultra} to featurize the entire image.
We train a Faster-RCNN~\cite{ren15faster} on \visualgenome{}~\cite{krishnavisualgenome}, with a ResNet101~\cite{resnet} backbone trained on JFT~\cite{hinton2015distilling} and fine-tuned on ImageNet~\cite{imagenet15}.
We select top-16 box proposals and featurize each of them with Graph-RISE, similar to \cite{changpinyo2019decoupled}.
Inspired by \cite{li20oscar}, we obtain up to 16 image tags from the Google Cloud Vision APIs, and treat them as text inputs to our model. These global, regional, and tag features end up being represented as a bag of 1+16+16 vectors, serving as bottom-up features \cite{anderson2018bottomup} for our model.

\mypar{Model and Learning}
For \ic{}-based pre-training and downstream tasks, we follow the state-of-the-art architecture that heavily rely on self-attention \cite{vaswani2017attention} or similar mechanisms~\cite{cc3m,yu2019multimodal,changpinyo2019decoupled,huang19attention,cornia20m2}. We implement a Transformer-based encoder-decoder model, using~\cite{changpinyo2019decoupled} as a starting point.
In addition, we encode each feature vector with a deeper embedding layer and apply layer normalization~\cite{ba2016layer}.
Following \cite{lu19vilbert}, we encode the corners and the area of bounding boxes and apply layer normalization when combining geometric and regional semantic features.
These modifications lead to an improved CIDEr score of 100.9 on the \cc{} dev benchmark (Table~\ref{tab:cc}), vs. 93.7 as reported by \cite{changpinyo2019decoupled}.
We describe additional details in the supplementary material, including infrastructure description, runtime, model size, hyperparameter ranges and tuning methods, and the configuration of the best-performing model.

For the \vlm{}-based pre-training and downstream tasks, we reuse the architecture above but discard the decoder.
We use mean pooling to obtain a fixed-length vector for each modality, and compute the product of the transformed (last-layer Transformer encoder representation) image and the transformed text before applying softmax.



\begin{table*}[t]
\small
\begin{center}
\begin{tabular}{l|cc|cc|cc|cc|}
& \multicolumn{8}{c}{\ttbf{nocaps val}} \\ \cline{2-9}
\multicolumn{1}{c|}{Method} & \multicolumn{2}{c|}{in-domain} & \multicolumn{2}{c|}{near-domain} &  \multicolumn{2}{c|}{out-of-domain} & \multicolumn{2}{c|}{overall}\\ \cline{2-9}
 & \cider & \spice & \cider & \spice & \cider & \spice & \cider & \spice \\ \hline
UpDown \cite{nocaps} & 78.1 & 11.6 & 57.7 & 10.3 & 31.3 & 8.3 & 55.3 & 10.1 \\
UpDown + CBS \cite{nocaps} & 80.0 & 12.0 & 73.6 & 11.3 & 66.4 & 9.7 & 73.1 & 11.1 \\
UpDown + ELMo + CBS \cite{nocaps} &79.3 & 12.4 & 73.8 & 11.4 & 71.7 & 9.9 & 74.3 & 11.2 \\ \hline
Oscar$_L$ \cite{li20oscar} & 79.9 & 12.4 & 68.2 & 11.8 & 45.1 & 9.4 & 65.2 & 11.4 \\
Oscar$_L$ + CBS \cite{li20oscar} & 78.8 & 12.2 & 78.9 & 12.1 & 77.4 & 10.5 & 78.6 & 11.8 \\
Oscar$_L$ + SCST + CBS \cite{li20oscar} & 85.4 & 11.9 & 84.0 & 11.7 & 80.3 & 10.0 & 83.4 & 11.4 \\ \hline
VIVO \cite{hu2020vivo} & 88.8 & 12.9 & 83.2 & \underline{12.6} & 71.1 & 10.6 & 81.5 & 12.2 \\
VIVO + CBS \cite{hu2020vivo} & 90.4 & \underline{13.0} & 84.9 & 12.5 & 83.0 & 10.7 & 85.3 & 12.2 \\
VIVO + SCST + CBS \cite{hu2020vivo} & \underline{92.2} & 12.9 & \underline{87.8} & \underline{12.6} & 87.5 & 11.5 & \underline{88.3} & \underline{12.4} \\ \hline
\emph{pretrain \ic{} on \ccp{}} & 88.3 & 12.3 & 86.0 & 11.8 & 91.3 & 11.2 & 87.4 & 11.8 \\
\emph{pretrain \ic{} on \cc{}+\ccp{}} & \textbf{92.6} & 12.5 & \textbf{88.3} & 12.1 & \underline{94.5} & \underline{11.9} & \textbf{90.2} & 12.1 \\ \hline
Human & 84.4 & \textbf{14.3} & 85.0 & \textbf{14.3} & \textbf{95.7} & \textbf{14.0} & 87.1 & \textbf{14.2} \\ \hline
& \multicolumn{8}{c}{\ttbf{nocaps test}} \\ \hline
UpDown \cite{nocaps}& 74.3 & 11.5 & 56.9 & 10.3 & 30.1 & 8.1 & 54.3 & 10.1 \\
UpDown + ELMo + CBS \cite{nocaps}& 76.0 & 11.8 & 74.2 & 11.5 & 66.7 & 9.7 & 73.1 & 11.2 \\ \hline
VIVO + SCST + CBS \cite{hu2020vivo} & \textbf{89.0} & \underline{12.9} & \textbf{87.8} & \underline{12.6} & 80.1 & 11.1 & \underline{86.6} & \underline{12.4} \\
\hline
\emph{pretrain \ic{} on \ccp{}} & 82.9 & 11.9 & 85.7 & 12.0 & 85.3 & 11.3 & 85.3 & 11.8 \\
\emph{pretrain \ic{} on \cc{}+\ccp{}} & \underline{87.2} & 12.3 & \underline{87.4} & 12.1 & \underline{87.2} & \underline{11.4} & \textbf{87.3} & 12.0 \\ \hline
Human & 80.6 & \textbf{15.0} & 84.6 & \textbf{14.7} & \textbf{91.6} & \textbf{14.2} & 85.3 & \textbf{14.7} \\ \hline
\end{tabular}
\vspace{-7pt}
\caption {\small Comparison between our best model (in \emph{italics}, pre-trained on \ccp{} with \ic{} and fine-tuned on \cococap{}) and existing models, on the \nocaps{} val (top) and test (bottom) splits. Bold indicates best-to-date, underline indicates second-best.}
\vspace{-25pt}
\label{tab:nocaps_sota}
\end{center}
\end{table*}

\begin{table}[t]
\small
\begin{center}
\begin{tabular}{l|c|c|}
 & \ttbf{COCO} & \ttbf{nocaps} \\
\multicolumn{1}{c|}{Method} & \ttbf{val2017} & \ttbf{val} \\ \cline{2-3}
 & \cider & \cider \\ \hline
UpDown (reference) & 116.2 & 55.3\\
UpDown + CBS & 97.7 & 73.1\\
UpDown + ELMo + CBS & 95.4 & 74.3\\ \hline
\emph{no pretrain (reference)} & 108.5 & 54.7 \\
\emph{pretrain \ic{} on \ccp{}} (5K) & 108.1 & \textbf{87.4} \\
\emph{pretrain \ic{} on \ccp{}} (10K) & \textbf{110.9} & 87.1 \\
\hline
\end{tabular}
\vspace{-7pt}
\caption {\small Performance on the in-domain \cococap{} val2017 split along with the \nocaps{} val split.
Our methods are in \emph{italics} with the number of fine-tuning steps in the parentheses.}
\vspace{-25pt}
\label{tab:coco}
\end{center}
\end{table}

\begin{table}[t]
\small
\begin{center}
\begin{tabular}{c|c|c|c|}
Pretraining & Finetuning & \ttbf{\locnarr{}} & \ttbf{\locnarr{}} \\
data & data & \ttbf{COCO val} & \ttbf{OID val} \\ \cline{3-4}
 &  & \cider & \cider \\ \hline
None & \locnarr{} COCO & 29.6 & 33.8 \\ \hline
\cc{} & \locnarr{} COCO & 29.1 & 35.7 \\
\ccp{} & \locnarr{} COCO & \textbf{30.0} & \textbf{38.6} \\ \hline
\end{tabular}
\vspace{-7pt}
\caption {\small Novel object captioning on \locnarr{}.}
\vspace{-20pt}
\label{tab:gen_locnar}
\end{center}
\end{table}

\begin{table}[t]
\small
\begin{center}
\begin{tabular}{l|c|c|}
& \ttbf{CC3M} & \ttbf{CC3M} \\
\multicolumn{1}{c|}{Method} & \ttbf{dev} & \ttbf{test} \\ \cline{2-3}
 & \cider & \cider \\ \hline
FRCNN \cite{changpinyo2019decoupled} & 89.2 & 94.4 \\
TTIC+BIU (single model) & - & 98.0 \\
Ultra~\cite{changpinyo2019decoupled} & 93.7 & 98.4 \\ \hline
\emph{no pretrain} & 100.9 & - \\
\emph{pretrain \ic{} on \ccp{} (no ft)} & 39.3 & -\\
\emph{pretrain \ic{} on \ccp{}} & \textbf{105.4} & -\\ \hline
\end{tabular}
\vspace{-7pt}
\caption {\small Performance on the Conceptual Captions (\cc{}) benchmark. Our methods are in \emph{italics}. ``ft" stands for fine-tuning. The top two \cc{} test CIDEr baseline scores are from the Conceptual Captions Leaderboard as of Nov 15, 2020.}
\vspace{-25pt}
\label{tab:cc}
\end{center}
\end{table}

\section{Experimental Results}
\label{sec:exp}

\subsection{Vision-to-Language Generation}

Table~\ref{tab:nocaps} shows our results on \textbf{\nocaps{}}.
We report in Row 1 the performance of our baseline model without pre-training.
Rows 2-3 show the performance of off-the-shelf captioning systems trained on \cc{} and \ccp{}, respectively.
This indicates the ``raw'' power (zero-shot setting) of the pre-trained network in generating captions out of the box.
We note that, without fine-tuning on \cococap{}, the model underperforms our baseline numbers on all metrics, which is indicative of the need for the model to learn the COCO captioning style, to which the existing automatic metrics are quite sensitive.
In addition, we observe a slightly better performance by \cc{} except for BLUE4 and SPICE. This illustrates the benefit of data processing and bias toward high-precision captions present in \cc{}.

With a fine-tuned model, the benefit of transfer learning using pre-training on this task is clear (Row 1 vs. Rows 4,5,6), with \ccp{} outperforming \cc{} by +14.2 CIDEr points and another +2.8 with \cc{}+\ccp{}.
Fig.~\ref{fig:nocaps_qual} illustrates this effect; 
scaling up pre-training data benefits learning multimodal correspondences from a much larger pool of concepts, potentially making the model less susceptible to hallucinations (e.g., guessing ``microphone'' as it has not seen ``bagpipes" in the training set), and also more informative (e.g. choosing ``sumo wrestlers'' over ``men''/``people'').

Table~\ref{tab:nocaps_sota} compares our best model (\ic{} pre-trained on \cc{}+\ccp{}) to existing state-of-the-art results on \nocaps{},
and show that ours achieves state-of-the-art performance on CIDEr, outperforming a concurrent work~\cite{hu2020vivo} that uses a different pre-training approach \emph{directly on the Open Images dataset, which \nocaps{} is based on.}
Importantly, we observe that the gain in the overall score can be largely attributed to the out-of-domain performance (3rd column). This result indicates that, although the annotation protocol for \nocaps{} uses the priming of annotators to mention one or more of displayed fine-grained ground-truth object classes (e.g., ``red panda'') present in the image~\cite{nocaps}, the large-scale and natural fine-grainedness of \ccp{} succeeds in correctly learning to generate captions containing such concepts, in spite of being textually out-of-domain.

Following \cite{nocaps}, we also report results of our best model on the \cococap{} val2017 split, see Table~\ref{tab:coco}, with 5K and 10K fine-tuning steps.
We note that, since we do not rely on techniques such as constrained beam search (CBS) \cite{anderson17guided,nocaps} that constrain the model outputs, we do not suffer from the large performance trade-offs seen with the previous solutions (degradation on in-domain performance as out-of-domain performance increases, see each model vs. ``reference").
Our result on out-of-domain data, as we vary the number of fine-tuning steps (last two rows), suggests that over--fine-tuning on \cococap{} may incur a cost in terms of poor generalization.

A second set of results is reported in Table~\ref{tab:gen_locnar}.
We observe that, even when the task requires the generation of much longer captions for \locnarr, \ccp{} achieves superior performance (as measured by CIDEr) compared to \cc{} as pretraining data.
However, the gain is smaller compared to the one observed for \nocaps{}.
We attribute this to the fact that injecting novel concepts into longer texts is harder, and also the fact that \locnarr{} does not use priming in their annotation process, leading to more generic terms in their annotation (``musical instruments" vs. ``trumpets").

Finally, we fine-tune our best pre-trained model (\ic{} on \ccp{}) using \cc{} in Table~\ref{tab:cc}, and then evaluate on the dev split.
We find that we improve the CIDEr score on the dev split from 100.9 to 105.4 (+4.5 CIDER points).
We note that the model trained on \ccp{} and evaluated directly on the \cc{} dev set (without fine-tuning on the \cc{} train split) obtains a low dev CIDEr of 39.3.
This again indicates that the additional processing steps done for \cc{} (e.g., hypernimization) result in captions that are different enough from the ones in \ccp{} to require a fine-tuning step. 


\begin{table}[t]
\small
\begin{center}
\begin{tabular}{c|c|ccc|}
Pretraining & Finetuning & \multicolumn{3}{c|}{\ttbf{\flickr{}}} \\ 
data & data & \multicolumn{3}{c|}{\footnotesize \ttbf{test}} \\
\cline{3-5}
 & & R1 & R5 & R10 \\ \hline
None & \flickr{} & 43.7 & 74.8 & 84.1 \\ \hline
\cc{} & None & 35.4 & 65.2 & 76.2 \\
\ccp{} & None & 42.5 & 73.1 & 83.4 \\ 
{\footnotesize \cc{}+\ccp{}} & None & \textbf{47.1} & \textbf{76.4} & \textbf{83.4} \\ \hline
\cc{} & \flickr{} & 52.3 & 81.7 & 88.4 \\
\ccp{} & \flickr{} & 58.5 & 86.6 & 92.1 \\ 
{\footnotesize \cc{}+\ccp{}} & \flickr{} & \textbf{61.5} & \textbf{87.5} & \textbf{92.8} \\ 
\hline \hline
Pretraining & Finetuning & \multicolumn{3}{c|}{\footnotesize \ttbf{\locnarr{} \flickr{}}} \\
data & data & \multicolumn{3}{c|}{\footnotesize \ttbf{test}} \\
\cline{3-5}
 & & R1 & R5 & R10 \\ \hline
None & {\footnotesize \locnarr{} \flickr{}} & 54.5 & 85.0 & 91.0 \\ \hline
\cc{} & {\footnotesize \locnarr{} \flickr{}} & 61.1 & 88.2 & 93.7 \\
\ccp{} & {\footnotesize \locnarr{} \flickr{}} & 70.2 & 92.1 & 95.6 \\
{\footnotesize \cc{}+\ccp{}} & {\footnotesize \locnarr{} \flickr{}} & \textbf{71.0} & \textbf{93.0} & \textbf{97.0} \\ \hline
\end{tabular}
\vspace{-7pt}
\caption {\small Image retrieval on \flickr{} and \locnarr{} \flickr{}}
\vspace{-25pt}
\label{tab:rtv_main}
\end{center}
\end{table}

\subsection{Vision-and-Language Matching}

Table~\ref{tab:rtv_main} reports zero-shot and default IR performance on \textbf{\flickr{}} as well as default IR performance on \textbf{\locnarr{} \flickr{}}.
The results are consistent with those in vision-to-language generation. First, both \cc{} and \ccp{} are beneficial, improving over ``from-scratch" training (Pretraining data as ``None") by at least 8.6\% and 6.6\% in R1 on \flickr{} and \locnarr{} \flickr{}, respectively.
Additionally, \ccp{} significantly outperforms \cc{} in all cases. Finally, combining the two datasets (\cc{}+\ccp{}) results in even better performance.
We provide qualitative results and additional discussion in the supplementary material.

Our zero-shot IR results (the three rows in Table~\ref{tab:rtv_main} with fine-tuning data as ``None'') are also competitive to the state-of-the-art, despite the fact that our model is much smaller (6 layers of transformers of hidden layer size 512 with 8 attention heads vs. 12 layers of size 768 with 12 attention heads) and uses late fusion instead of early fusion.
In particular, our zero-shot IR on \cc{} outperforms the one in ViLBERT \cite{lu19vilbert} (35.4 vs. 31.9 in R1),
while the \ccp{} performance goes up by +7.1\% R1 to 42.5, and an additional +4.6\% R1 to 47.1 when using \cc{}+\ccp{}, surpassing the ``from-scratch" setting.

\section{Related Work}
\label{sec:related}

\mypartop{V+L Pre-training}
V+L pre-training research makes use existing large-scale datasets with image-text pairs.
A majority of these resources are image captioning datasets.
\cc{}~\cite{cc3m} has been the most popular for pre-training~\cite{lu19vilbert,lu2012in1,alberti19fusion,su20vlbert,zhou20unified,li2020unicoder,chen20uniter,li20oscar}.
Smaller but less noisy \sbu{}~\cite{sbucap} (\~1M) and \cococap{}~\cite{cococap} (106K) datasets are also of high interest.
Some work ~\cite{tan19lxmert,chen20uniter,li20oscar} use V+L resources collected for dense captioning or visual question answering (VQA), such as VG~\cite{krishna2017visual}, VQA2~\cite{goyal2017making}, and GQA~\cite{hudson2019gqa}. 
In contrast, \ccp{} is not collected for specific target tasks, and thus it is order-of-magnitude larger than those datasets.\footnote{Recently appearing after we submitted our paper, ALIGN~\cite{jia2021scaling}, CLIP~\cite{radford2021learning}, WIT~\cite{srinivasan2021wit}, WenLan~\cite{huo2021wenlan} all explore enlarging Web-scale data for V+L pre-training with success (albeit with different focuses), further confirming our intuition that scale is a critical factor.}
Furthermore, it is much more visually diverse, especially given the fact that \cococap{}, VG, VQA2, GQA are built on top of COCO images~\cite{coco} or its subsets.

Objectives in V+L pre-training research are largely influenced by BERT~\cite{devlin19bert}.
Masked language modeling has been extended to visual region inputs, while the next sentence prediction is analogous to \vlm{}. Based directly upon BERT, V+L pre-training research has largely been focused on V+L \emph{understanding}~\cite{lu19vilbert,li19visualbert,chen20uniter, tan19lxmert,alberti19fusion,su20vlbert, li2020unicoder,lu2012in1}, with classification or regression tasks that do not involve generation.
One exception is UnifiedVL~\cite{zhou20unified}, which pre-trains a unified architecture for both image captioning (generation) and VQA (understanding). Our work focuses on simpler objectives and consider one at a time. This allows for a ``clean" study of the effect of pre-training data sources. At the same time, we also pre-train vision-to-language generation and encoder-decoder jointly as opposed to an encoder-only setup.
Our work also shows that \ic{} is a strong objective for vision-to-language generation with respect to the widely-used masking-based objectives. Consistent with our results, \ic{} is successfully adopted for learning visual representations for lower-level vision tasks~\cite{desai2021virtex}.

\mypar{Long-tail Visual Recognition in V+L}
Addressing long-tail distributions of visual concepts is an important component of V+L systems that generalize, as long and free-form texts exhibit a large number of compositional, fine-grained categories \cite{zhu2014capturing,liu2019large,changpinyo2019decoupled}. Our work focuses on downstream testbeds for V+L research that require this adaptation ability. For example, the train-test distribution discrepancy in \nocaps{} exists in both visual (\coco{} vs. \oid{}) and textual domains (80 to object classes vs. 600 classes). The same can be said for zero-shot image retrieval \cite{lu19vilbert}, in which the model must generalize visually and textually from the pre-training data sources of \cc{} or \ccp{} to Flickr30K. Our work identifies pre-training with large-scale noisy data as a promising solution. In addition, for the task noval object captioning, our approach works more robustly across in- and out-of- domain scenarios and is simpler than the state-of-the-art techniques that utilize constrained beam search (CBS) \cite{anderson17guided}, finite state machine construction plus CBS \cite{anderson18partially}, generating slot-filling templates \cite{lu18cvpr,wu2018decoupled}, and copying mechanisms \cite{yao17incorporating}.


\section{Conclusion}
\label{sec:discuss}

We introduce the new V+L pre-training resource \ccp{}, obtained by extending the pipeline in \cite{cc3m}. 
We show that the scale and diversity of V+L pre-training matters on both generation and matching, especially on benchmarks that require long-tail recognition such as \nocaps{}. Our results indicate leveraging noisy Web-scale image-text pairs as a promising direction for V+L research.

{\small
\mypar{Acknowledgments}
We thank Peter Anderson for his feedback on earlier version of the draft, Bo Pang, Zhenhai Zhu for helpful discussions, Sebastian Goodman and Ashish V. Thapliyal for help with model implementation, Chris Alberti for help with the data collection pipeline, and Harsh Agrawal for detail on \nocaps{}.}

{\small
\bibliography{arxiv_main}
\bibliographystyle{ieee_fullname}
}

\appendix


\section{Broader Impact}
\label{sec:broader}

Our publicly-available V+L pre-training resource \ccp{} has the potential to positively impact multiple vision-and-language tasks. One main aspect that we have identified is a much higher degree of coverage of long-tail visual concepts than previous resources, including \cc{}. As a result, we expect the models (pre-)trained on our data to be more robust in the wild than before.

In addition, our work could benefit the design of new setups for the downstream tasks that shift away from in-domain (e.g., COCO/Visual Genome) to out-of-domain/in-the-wild (e.g., OID), similar to \nocaps{} that our work focuses heavily on. The setups could also avoid the use of in-domain data during \emph{pre-training} that in some cases resulting in transfer learning between (almost) identical sets of images, e.g., COCO, Visual Genome (VG), VQA2, VG QA, Visual7W, GQA, GuessWhat, and RefCOCO*. 

At the same time, datasets curated from the Web could come with risks such as unsuitable content (adult content, profanity) and unintended privacy leakage \cite{nasr2019comprehensive,carlini2019secret,carlini2020extracting}. We take the steps in Sect.~2.2 of the main text to mitigate both of these risks by applying the necessary image and text filtering steps and replacing each person name (celebrities' included) with the special $<$PERSON$>$ token.

Less specific to the Web data are the unwanted dataset biases~\cite{burns2018women,stock2018convnets,wang2019balanced} that are prone to amplification by machine learning models~\cite{bolukbasi2016man,zhao2017men}. Our preliminary analysis in Sect.~2.3 of the main text shed light on the degree to which our data exhibits some aspects of these inherent biases, and we suspect that the better coverage of the tail in fact makes this issue less severe. Nevertheless, the users of this data and the systems trained on it shall be aware of such risks and other ones that might arise. 


\section{Additional analyses of \ccp{}}

\subsection{Out-of-domain (OOD) visual concepts on an expanded list of datasets}
We use the 394 nocaps’ out-of-domain classes as a proxy for OOD visual concepts and analyze popular vision-and-language datasets, in addition to \cc{} and \ccp{} that we focus in the main text.
These datasets span a wide range of use cases, both in terms of tasks (image-to-text generation, image-and-text matching, visual question answering (VQA), referring expression comprehension, and multimodal verification), and in terms of the stage during which they are used (pre-training, fine-tuning/evaluation, or both.)

\begin{itemize}[noitemsep,topsep=0pt,parsep=0pt,partopsep=0pt]
  \item \cc{} \cite{cc3m} An instance of text is the caption associated with each image url of the training split.
  \item \ccp{} (ours) An instance of text is the caption associated with each image url. It has been used and is currently the most popular V+L pre-training dataset~\cite{lu19vilbert,alberti19fusion,chen20uniter,su20vlbert,zhou20unified,lu2012in1,li20oscar}.
  \item \cococap{} \cite{cococap} An instance of text comes from the caption associated with each image of the 2017 training split (five captions per image). This dataset is designed for the task of image captioning, and has been used for caption-based image retrieval as well. It has been used for V+L pre-training~\cite{tan19lxmert,li19visualbert,chen20uniter,li20oscar}.
  \item \visualgenome{} \cite{krishnavisualgenome} An instance of text comes from the caption of each region in images of the training split. This dataset aims to connect vision and language through scene graphs and is used for multiple tasks that include but not limited to dense image captioning, visual relationship detection and scene graph parsing, image retrieval and generation, and visual question answering. It has been used for V+L pre-training~\cite{tan19lxmert,chen20uniter}.
  \item \sbu{} \cite{sbucap} An instance of text is the caption associated with each image url of the ``preferred" version of the dataset. This dataset is designed for the task of image captioning. It has been used for V+L pre-training~\cite{tan19lxmert,chen20uniter,li2020unicoder,li20oscar}.
  \item \vqatwo{} \cite{goyal2017making} An instance of text is the question and the answers in each image-question-answers triplet of the train2014 + val2train2014 splits. This dataset is designed for the task of visual question answering (VQA)~\cite{antol2015vqa}. It has been used for V+L pre-training~\cite{tan19lxmert,li20oscar}.
  \item \cocorefg{} \cite{refcocog} An instance of text is the referring expression in each region in images of the training split. This dataset is designed for the task of referring expression comprehension~\cite{refcoco}.
  \item \nlvrtwo{} \cite{nlvr2} An instance of text comes from the caption associated with each pair of images of the training split. This dataset is used for the task called multimodal verification in \cite{lu2012in1}, but designed for the general task of visual reasoning.
\end{itemize}

Table~\ref{tab:nocaps_ood} summarizes the number of instances whose texts contain OOD visual concepts for all selected datasets. We use both the absolute frequency and the normalized one (per 1M text instances). Essentially, these numbers indicate the degree of OOD coverage. We find that \ccp{} has many more OOD instances than all other datasets by a large margin (6.7x median and 5.8x mean vs. the second best \cc{}). Moreover, \ccp{} still prevails \emph{even after normalization} to account for its size. In other words, \ccp{} covers these OOD classes better in both absolute and relative senses.

\begin{figure}[t]
\begin{center}
\vspace{-12pt}
\includegraphics[width=.8\linewidth]{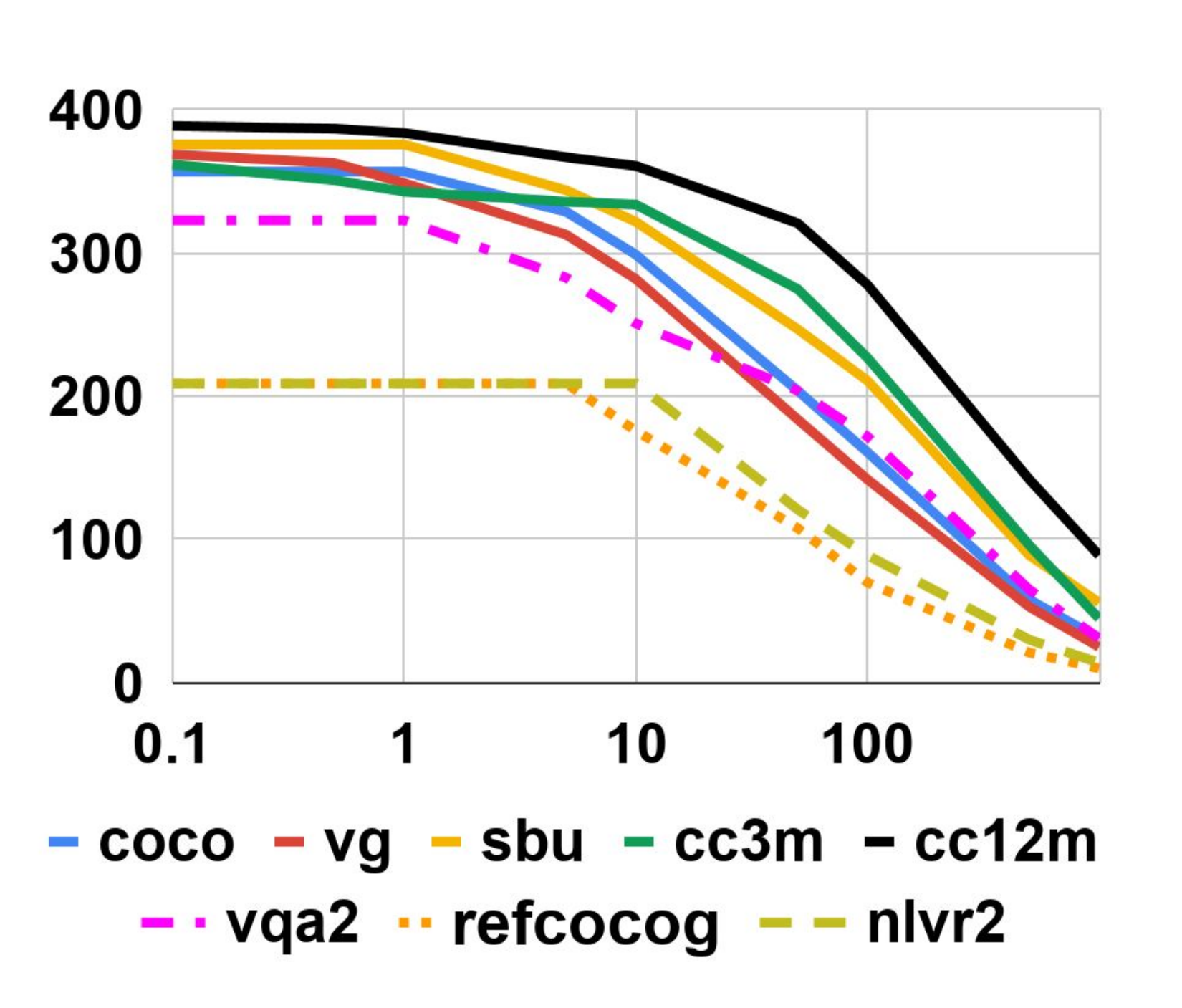}
\end{center}
 \vspace{-15pt}
 \caption{\small Comparison of nocaps' out-of-domain coverage degree among captioning (solid) and 3 other tasks' (dashed) datasets (see text for details).}
\label{fig:nocaps_ood}
\vspace{-8pt}
\end{figure}

Fig.~\ref{fig:nocaps_ood} provides a more complete picture of the normalized frequency of OOD classes in these datasets, at different thresholds. It shows the number of OOD classes (y-axis) with at least $K$ per 1M captions (x-axis). Evidently, other datasets experience sharper drops as $K$ increases than \ccp{} (black solid curve). We also find that captioning datasets (solid curves) generally provide better coverage than non-captioning datasets: \vqatwo{}, \cocorefg{}, and \nlvrtwo{} (dashed curves).

\begin{table}[t]
\small
\begin{center}
\begin{tabular}{l|cc|cc|}
\multicolumn{1}{c|}{Dataset} & \multicolumn{2}{c|}{Freq} & \multicolumn{2}{c}{Freq (per 1M)} \\ \cline{2-5}
 & median & mean & median & mean \\ \hline
\cc{} & 462 & 2325.7 & 139.2 & 700.8 \\
\ccp{} & \textbf{3110} & \textbf{13455.8} & \textbf{250.3} & \textbf{1083.1} \\ \hline
\cococap{} & 37 & 248.6 & 62.3 & 417.1 \\
\visualgenome{} & 133 & 1114.47 & 40.7 & 341.4 \\
\sbu{} & 121 & 798.6 & 121.0 & 798.6 \\ \hline \hline
\vqatwo{} & 37 & 242.0 & 63.8 & 417.2 \\
\cocorefg{} & 1 & 21.2 & 8.8 & 186.4 \\
\nlvrtwo{} & 4 & 79.9 & 11.6 & 245.5 \\ \hline
\end{tabular}
\vspace{-7pt}
\caption {\small Statistics of the (normalized) frequency of \nocaps{}' out-of-domain visual concepts in the texts of popular vision-and-language datasets.}
\label{tab:nocaps_ood}
\vspace{-10pt}
\end{center}
\end{table}

\begin{figure}[t]
\begin{center}
\includegraphics[width=.48\linewidth]{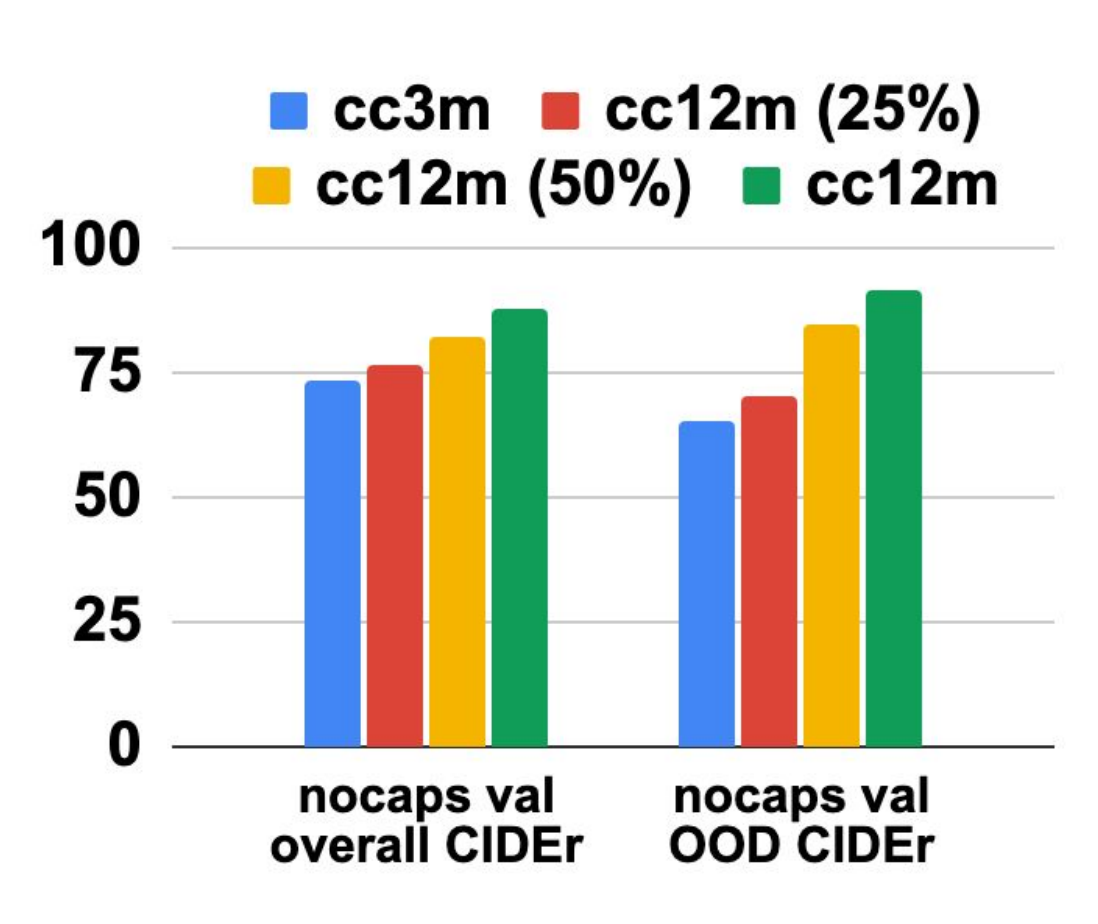}
\includegraphics[width=.48\linewidth]{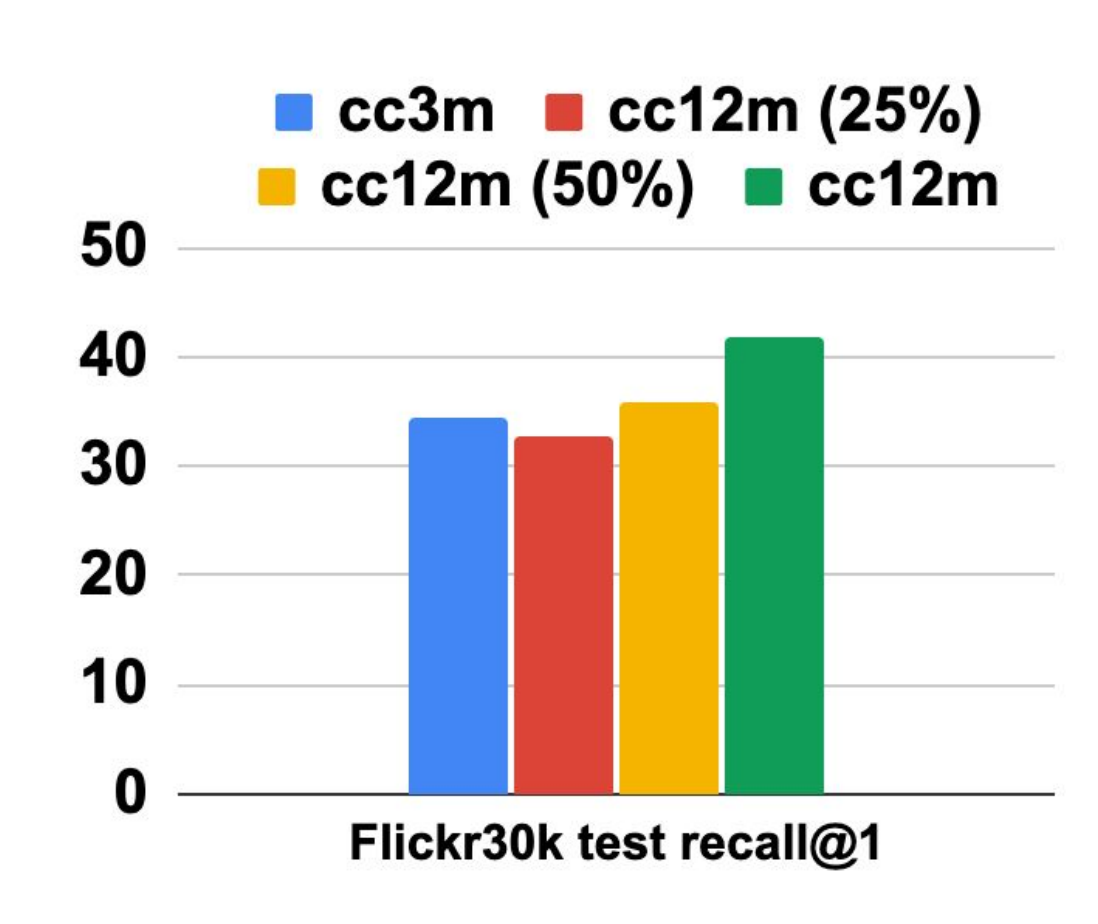}
\end{center}
 \vspace{-15pt}
 \caption{\small Performance with sub-sampled \ccp{} (25\% \& 50\%) on novel object captioning (left, CIDEr's on \nocaps{} val) and zero-shot IR (right, recall@1 on \flickr{} test).}
\label{fig:cc12m_partial}
\vspace{-15pt}
\end{figure}

\subsection{The impact of the dataset size}
We experiment with pre-training on randomly subsampled \ccp{}, 25\% (3.1M) and 50\% (6.2M) and evaluate the pre-trained models on novel object captioning on \nocaps{} and zero-shot IR on \flickr{}. Fig.~\ref{fig:cc12m_partial} shows the larger, the better trend, with 25\% of \ccp{} gives rise to similar performance as CC3M.

\begin{figure*}[t]
\begin{center}
\includegraphics[width=0.33\linewidth]{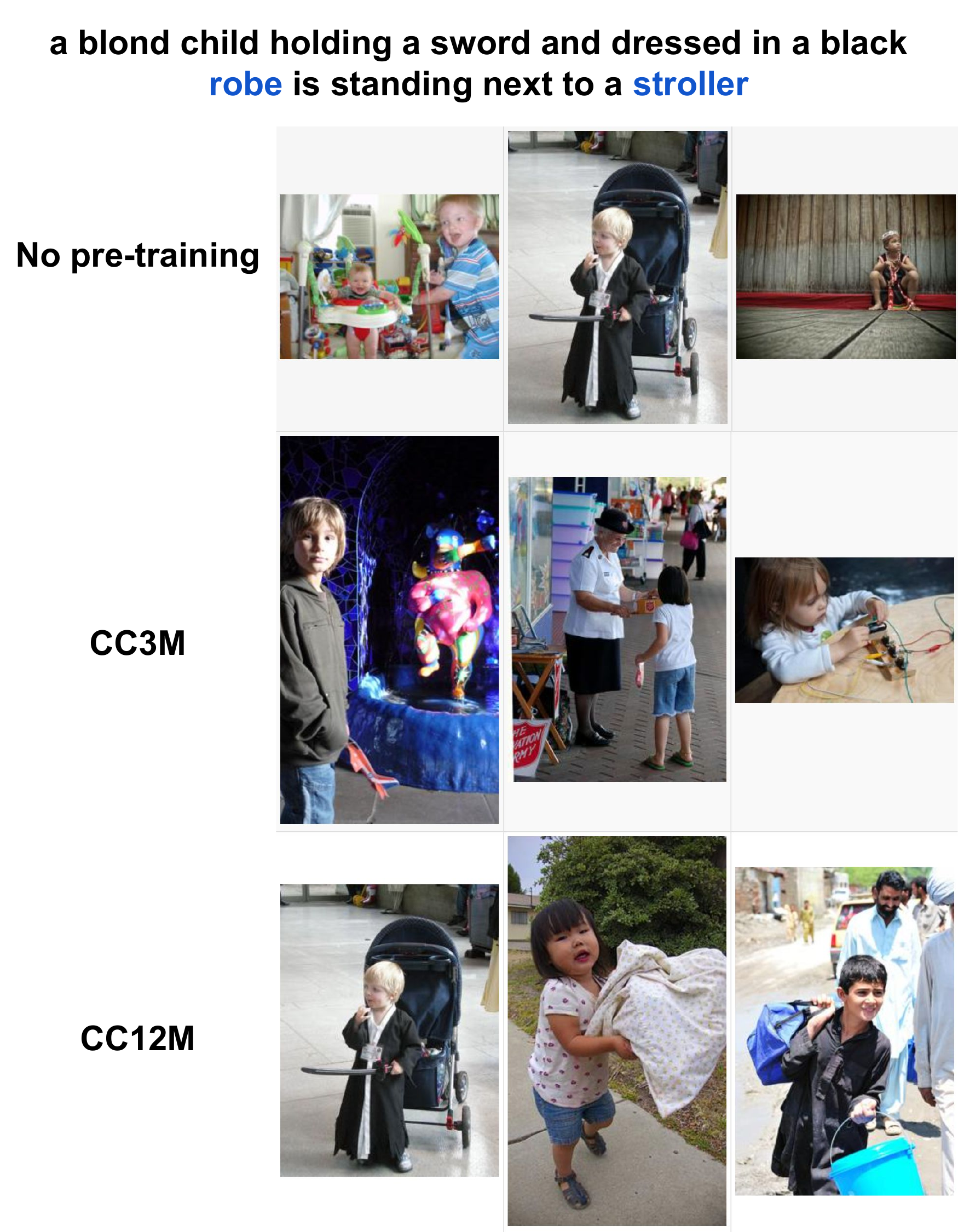}
\includegraphics[width=0.33\linewidth]{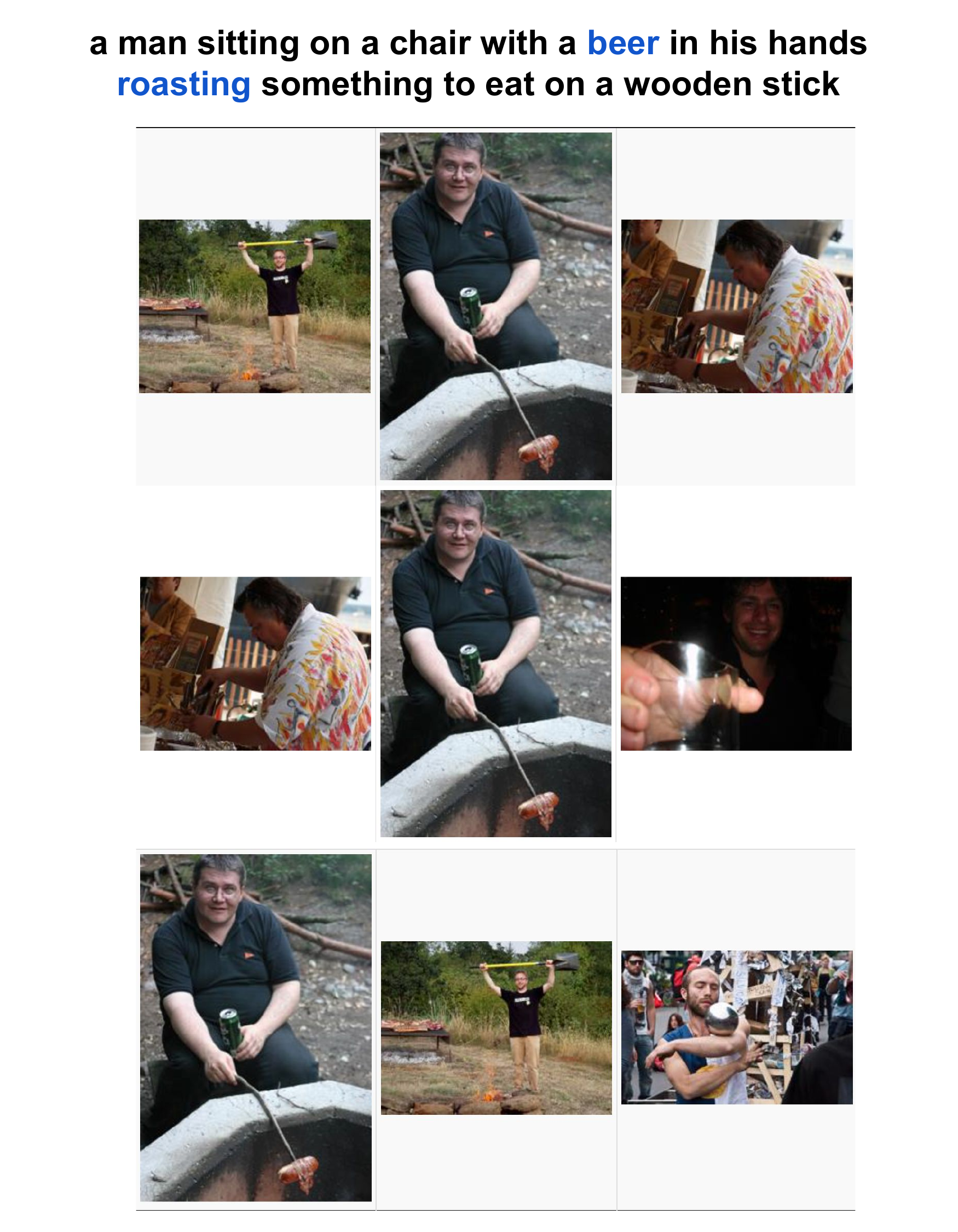}
\includegraphics[width=0.33\linewidth]{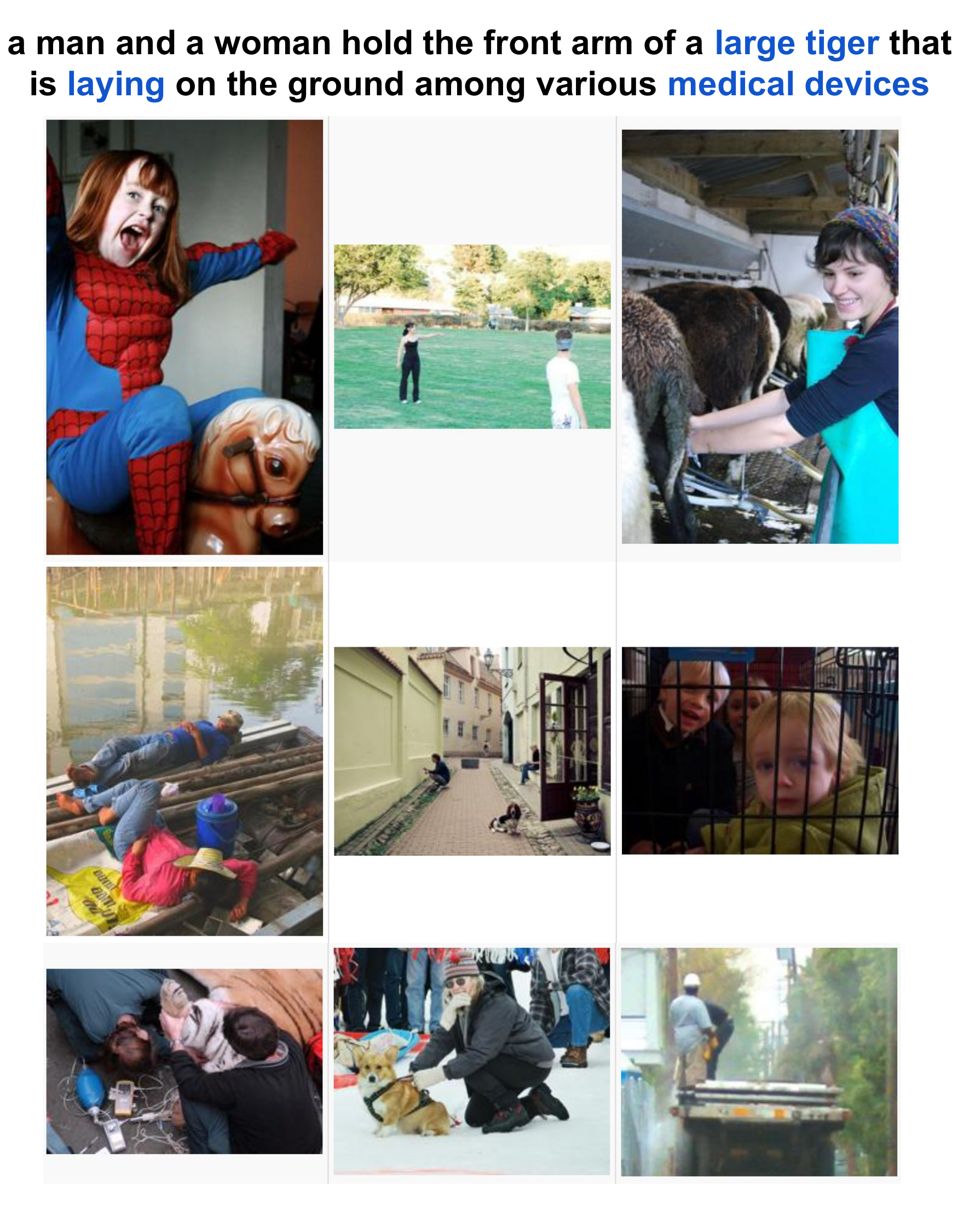}
\end{center}
 \vspace{-12pt}
 \caption{\small \textbf{Qualitative results for the image retrieval task} on \flickr{} given the query text (very top) when the model is not pre-trained (top), pre-trained on \cc{} (middle), and pre-trained on \ccp{} (bottom).}
\label{fig:flickr_qual}
\vspace{-15pt}
\end{figure*}

\section{Qualitive Results for Image Retrieval}
Fig.~\ref{fig:flickr_qual} provides qualitative image retrieval results on the \flickr{} dataset, top-3 images retrieved by the from-scratch model trained on \flickr{}, as well as by two models pre-trained on \cc{} and \ccp{} and then fine-tuned on \flickr{}. We report three cases in which \ccp{} pre-training helps correct the rankings from the other two models, which we suspect due to the model getting more familiar with the rare words, highlighted in blue.


\begin{table*}[t]
\small
\begin{center}
\begin{tabular}{r|cc|cc|cc|cccccc|}
\multicolumn{1}{c|}{} & \multicolumn{12}{c}{\ttbf{nocaps val}} \\ \cline{2-13}
\multicolumn{1}{c|}{Pre-training} & \multicolumn{2}{c|}{in-domain} & \multicolumn{2}{c|}{near-domain} &  \multicolumn{2}{c|}{out-of-domain} & \multicolumn{6}{c|}{overall}\\ \cline{3-13}
 \multicolumn{1}{c|}{data} & \cider & \spice & \cider & \spice & \cider & \spice & \bleuo & \bleuf & \meteor & \rouge & \cider & \spice \\ \hline
 \locnarr{} \oid{} & 76.0 & 11.6 & 65.9 & 10.9 & 48.9 & 9.3 & 73.3 & 17.4 & 23.5 & 50.7 & 63.9 & 10.7 \\
 \cc{} & 81.8 & 11.6 & 73.7 & 11.1 & 65.3 & 10.1 & 74.6 & 19.1 & 24.1 & 51.5 & 73.2 & 11.0 \\
\ccp{} & 88.3 & 12.3 & 86.0 & 11.8 & 91.3 & 11.2 & 78.5 & 23.4 & 25.9 & 54.5 & 87.4 & 11.8 \\
 \hline
\end{tabular}
\vspace{-1pt}
\caption {Comparison between pre-training data. \locnarr{} \oid{}'s images are from the same visual domain as \nocaps{}. All approaches use the \ic{} pre-training objective.}
\vspace{-1pt}
\label{tab:nocaps_ptdata}
\end{center}
\end{table*}

\begin{table*}[t]
\small
\begin{center}
\begin{tabular}{r|cc|cc|cc|cccccc|}
\multicolumn{1}{c|}{} & \multicolumn{12}{c}{\ttbf{nocaps val}} \\ \cline{2-13}
\multicolumn{1}{c|}{Pre-training} & \multicolumn{2}{c|}{in-domain} & \multicolumn{2}{c|}{near-domain} &  \multicolumn{2}{c|}{out-of-domain} & \multicolumn{6}{c|}{overall}\\ \cline{2-13}
 \multicolumn{1}{c|}{objective} & \cider & \spice & \cider & \spice & \cider & \spice & \bleuo & \bleuf & \meteor & \rouge & \cider & \spice \\ \hline
 \ic{} & 88.3 & 12.3 & 86.0 & 11.8 & 91.3 & 11.2 & 78.5 & 23.4 & 25.9 & 54.5 & 87.4 & 11.8 \\ \hline
 \mlmrate{.1} & 76.4 & 11.5 & 68.4 & 10.8 & 57.6 & 9.6 & 73.0 & 18.1 & 23.5 & 50.6 & 67.4 & 10.6 \\
 \mlmrate{.2} & 79.8 & 11.3 & 76.3 & 10.9 & 76.2 & 10.2 & 76.2 & 20.5 & 24.1 & 52.4 & 76.8 & 10.8 \\
 \mlmrate{.4} & 86.5 & 12.3 & 82.7 & 11.5 & 86.3 & 11.3 & 78.0 & 22.7 & 25.2 & 53.7 & 84.0 & 11.6 \\
 \mlmrate{.8} & 89.3 & 12.5 & 87.5 & 11.9 & 91.1 & 11.3 & 78.7 & 23.8 & 25.9 & 54.4 & 88.5 & 11.9 \\
 \hline
 \massrate{.1} & 86.0 & 12.1 & 74.8 & 11.1 & 71.7 & 10.1 & 75.8 & 20.5 & 24.6 & 52.5 & 75.8 & 11.0 \\
 \massrate{.2} & 84.9 & 12.0 & 78.1 & 11.2 & 78.6 & 10.5 & 76.0 & 20.8 & 24.7 & 52.7 & 79.2 & 11.2 \\
 \massrate{.4} & 85.7 & 11.7 & 83.7 & 11.5 & 88.5 & 10.9 & 77.3 & 22.8 & 25.1 & 53.6 & 85.0 & 11.4 \\
 \massrate{.8} & 88.8 & 12.2 & 85.1 & 11.7 & 87.8 & 10.6 & 78.1 & 23.7 & 25.5 & 54.2 & 86.2 & 11.5 \\
 \hline
 
\end{tabular}
\vspace{-6pt}
\caption {Comparison between the \ic{} pre-training and masked V+L pre-training. We consider two masking schemes (\mlm{} and \mass{}) and four masking rates (.1, .2, .4, .8) and report their effects on the \nocaps{} val set.}
\vspace{-12pt}
\label{tab:nocaps_masking}
\end{center}
\end{table*}

\begin{table*}[t]
\small
\begin{center}
\begin{tabular}{r|cc|cc|cc|cccccc|}
\multicolumn{1}{c|}{} & \multicolumn{12}{c}{\ttbf{nocaps val}} \\ \cline{2-13}
\multicolumn{1}{c|}{Pre-training} & \multicolumn{2}{c|}{in-domain} & \multicolumn{2}{c|}{near-domain} &  \multicolumn{2}{c|}{out-of-domain} & \multicolumn{6}{c|}{overall}\\ \cline{3-13}
 \multicolumn{1}{c|}{objectives} & \cider & \spice & \cider & \spice & \cider & \spice & \bleuo & \bleuf & \meteor & \rouge & \cider & \spice \\ \hline
 \ic{} & 88.3 & 12.3 & 86.0 & 11.8 & 91.3 & 11.2 & 78.5 & 23.4 & 25.9 & 54.5 & 87.4 & 11.8 \\
 \ic{}+\vlm & 88.6 & 12.3 & 85.8 & 11.9 & 90.0 & 11.4 & 78.0 & 23.1 & 25.7 & 54.4 & 87.1 & 11.9 \\
 \ic{}+\moc & 91.1 & 12.4 & 88.4 & 12.1 & 93.6 & 11.4 & 78.8 & 24.6 & 26.2 & 55.2 & 89.9 & 12.0 \\
 \hline
\end{tabular}
\vspace{-1pt}
\caption {Effect of visual linguistic matching (\vlm{}) and masked object classication (\moc{}) when combined with the \ic{} objective on the \nocaps{} val set.}
\vspace{-1pt}
\label{tab:nocaps_vlmmoc}
\end{center}
\end{table*}

\section{Pre-Training: Data and Method Variants}

\subsection{Vision-to-Language Pre-Training on \locnarr{} \oid{}}
Table~\ref{tab:nocaps_ptdata} considers pre-training on \locnarr{} \oid{} for the \nocaps benchmark. We observe inferior performance to both \cc{} and \ccp{}. 
We attribute this to the long narratives in \locnarr{} having drastically different styles from those from \cococap{} and \nocaps{}. Furthermore, the data collection protocol in \nocaps{} does not involve priming the annotator to mention object names present to the user, resulting in more generatic terms (instrument vs. guitar). This again highlights the natural fine-grainedness inherent in noisy Web data, especially in the case of no-hypernymized data source (\ccp{}).

\subsection{Pre-Training Strategies}

In the main text, we focus on the image captioning (\ic{}) and the visual-linguistic matching (\vlm{}) learning objectives both during pre-training and fine-tuning stages. Our motivation here is to keep the setup for evaluating pre-training data as ``clean" as possible. However, other pre-training strategies exist in the literature and we describe and test the effectiveness of them in this section. 

\subsubsection{Masked Vision-to-Language Generation}

Given the training image-text pairs, the \ic{} objective predicts the text from the image. The following objectives predict (all or part of) the text from the image \emph{and} (all or part of) the text. In order to \emph{encode} both the image and the text, we concatenate the sequence of image feature vectors and the sequence of text token feature vectors, and use the Transformer encoder to encode them \cite{li19visualbert,chen20uniter,su20vlbert}. This vanilla fusion is effective, shown to consistently outperform the co-attentional transformer layer \cite{lu19vilbert,lu2012in1}, in which the ``query'' comes from the other modality than that of ``key'' and ``value'' (see Sect.~2 and Fig.~2 in \cite{lu19vilbert} for details).

\mypar{Masked Language Modeling (\mlm)}
We mask a percentage of the input text tokens at random, and predict the target text sequence using the decoder.
Following BERT \cite{devlin19bert}, we use a mixed strategy for masking: for each selected token, we replace it with the mask token [MASK] 80\% of the time, replace it with a random token 10\% of the time, and leave it as is 10\% of the time.

\mypar{Masked Sequence to Sequence Modeling (\mass{})}
We apply the mixed masking strategy as in \mlm{} to the input text tokens, but require that the mask is applied to consecutive tokens (i.e., a contiguous segment).
The task is to sequentially predict the masked segment using the decoder.
This approach is inspired by MASS~\cite{song19mass} and PEGASUS~\cite{zhang20pegasus}.

\mypar{Results}
Table~\ref{tab:nocaps_masking} compares \ic{}, \mlm{}, and \mass{} pre-training objectives. Our main observation is that \ic{} clearly outperforms masked vision-to-language pre-training when the masking rate is low. Overall, \ic{} is competitive to \mlm{} and \mass{}, slightly below \mlmrate{.8} in overall CIDEr, but higher on out-of-domain CIDEr.

In addition, the trend suggests that it is critical that the text masking rate is high enough such that the models become less and less reliant on text --- that is, when \mlm{} and \mass{} become more similar to the \ic{} task.
Note that widely-used configurations in the VLP literature on vision-and-language understanding are the ones with low text masking rates (0.2 in most cases), which consistently underperform in our generation setup.

We attribute this result to the models' (over)reliance on text during pre-training, which hurts the quality of its \emph{image} representations. Supporting evidence for this phenomenon is found in the recent work of \cite{cao20behind}, which observe that image+text pre-trained models exhibit a preference for attending text rather than images during inference (in image and text understanding task). Another supporting evidence is the issue of strong language priors (well-known in the VQA community), which led to interest in \emph{adversarial} test sets and other methods to overcome strong language biases~\cite{vqacp,ramakrishnan18overcoming,clark19don,cadene19rubi}. The same pheonmenon has been reported for multi-modal machine translation, where models trained on image+text tend to ignore the image and primarily use the text input~\cite{caglayan2019probing}. Based on these results, the design of V+L pre-training objectives that are capable of outperforming the image-only \ic{} objective (i.e., overcoming the language  through modeling) is an interesting venue for future work.

Another observation is that \mass{} significantly works better than \mlm{} for lower masking rates. When masking rates are high, the two objectives become more similar. This suggests the importance of bridging the gap between pre-training and fine-tuning (producing consecutive tokens). 

\subsubsection{Image Captioning with Visual-Linguistic Matching or Masked Object Classification}

We explore adding auxiliary losses to the main \ic{} objective. First, we define a pre-training task that does not require text.

\mypar{Masked object classification (\moc{})}
We mask one of the visual regions (selected at random), and predict the cluster ID of that region~\cite{lu19vilbert,tan19lxmert,chen20uniter}.
We use a total of 8192 clusters, obtained via K-means over the training data.

Then, we either add the \vlm{} loss (multipled by 0.1) or the \moc{} loss (multipled by 0.1) to the main \ic{} loss. 

\mypar{Results}
Table~\ref{tab:nocaps_vlmmoc} reports the effect of multi-task pre-training on the \nocaps{} val set. We observe a slight improvement when adding \moc{} but a slight drop when adding \vlm{}. This again shows that \ic{} is a good pre-training task to start with. We leave developing advanced auxiliary losses on top of it and multi-task pre-training strategies for future work.


\section{Implementation Details}

\subsection{Data Preprocessing and Feature Embedding}

\begin{itemize}[noitemsep,topsep=0pt,parsep=0pt,partopsep=0pt]
  \item Text tokenizer: preprocesed with COCO tokenizer  \url{https://github.com/tylin/coco-caption}. We then create a vocabulary of subtokens out of these.
  \item Text input embedding (during pre-training only): subtoken lookup embeddings of size E = 512 are randomly initialized, followed by Linear(512)-ReLU-Dropout(0.3)-Linear(512).
  \item Image's geometric features: two pairs of coordinates (top left and bottom right) and the relative area, represented by \emph{relative} numbers between 0 and 1. Each of these 5 numbers is linearly projected into an embedding of size E = 512. We concatenate the result to get an embedding of size E x 5 = 2560, followed by Linear(512)-ReLU-Dropout(0.3)-Linear(512).
  \item Image's semantic features: each feature vector (a global image feature vector or one of the 16 box's image feature vector, followed by Linear(512)-ReLU-Dropout(0.3)-Linear(512).
  \item Image's combined geometric and semantic features: we first apply LayerNorm \cite{ba2016layer} to each of the geometric or the semantic features. We then add the two and apply Linear(512)-ReLU-Dropout(0.3)-Linear(512)-LayerNorm.
  \item Image's tag features: same as text input embedding.
\end{itemize}
For the \ic{} objective, we have a bag of 1 + 16 visual feature vectors and up to 16 tag feature vectors, each of size 512. For the \vlm{} objective, where text has to be encoded, we also have a sequence of text (sub)token feature vectors of size 512.

\subsection{Model}
The \ic{}-based task uses a transformer encoder-decoder model. The \vlm{}-based uses two transformer encoders, one for texts and the other for images.

\begin{itemize}[noitemsep,topsep=0pt,parsep=0pt,partopsep=0pt]
  \item Transformer image encoder: number of layers L = 6.
  \item Transformer image encoder: vocab embedding size E = 512.
  \item Transformer image encoder: hidden embedding size H = 1024.
  \item Transformer image encoder: feedforward/filter size F = H x 4 = 4096, following \cite{devlin19bert}.
  \item Transformer image encoder: number of attention heads A = H / 64 = 8, following \cite{devlin19bert}.
  \item Transformer text encoder (for \vlm{} only): L, E, H, F, A are the same as Transformer image encoder.
  \item Transformer decoder: L, E, H, F, A are the same as Transformer image encoder.
  \item Transformer decoder: beam search width = 5.
  \item Transformer decoder: beam search alpha = 0.6.
  \item Transformer decoder: maximum output length = 36 for all datasets except for \locnarr{} which is set to 180.
\end{itemize}

\subsection{Training}
\begin{itemize}[noitemsep,topsep=0pt,parsep=0pt,partopsep=0pt]
  \item Infrastructure: Google Cloud 32-core TPUs.
  \item Batch size per core: 128 (for a total of 4096)
  \item Optimizer: Adam \cite{adam} with default hyperparameters (except for the initial learning rate; see below).
  \item Learning rate --- Initial: See Hyperparameter search below. 
  \item Learning rate --- Warm-up epochs: 20 for all pre-training and fine-tuning experiments.
  \item Learning rate --- Decay rate: 0.95 for all pre-training and fine-tuning experiments.
  \item Learning rate --- Decay epochs: 25 for all pre-training and fine-tuning experiments.
  \item Data augmentation: a set of input visual regions are permuted during training.
  \item Maximum number of steps: 2M for vision-to-language generation pre-training on both \ccp{} and \cc{} (and \cc{}+\ccp{}). For vision-and-language matching, 1M for \cc{} instead. See Hyperparameter search below for fine-tuning experiments.
\end{itemize}

\subsection{Evaluation}

For \nocaps{} evaluation, we submit inference results to the leaderboard \url{https://evalai.cloudcv.org/web/challenges/challenge-page/464/overview}. Code for all evaluation metrics can be found at \url{https://github.com/nocaps-org/updown-baseline/blob/master/updown/utils/evalai.py}. For in-depth discussions of these metrics see \cite{kilickaya17reevaluating}.

Participating in the default formulation of the \nocaps{} challenge requires that one (i) does not use val and test \oid{}'s ground-truth object detection annotations, and (ii) does not use image-caption data collected via additional annotation protocols.
We satisfy both requirements as we train our object detector on Visual Genome, and both \cc{} and \ccp{} are automatically harvested from the web (alt-text) and belong to the category of noisy web data, therefore satisfying the second requirement.
On the other hand, models that leverage the Open Images Localized Narratives dataset (\locnarr{}) \cite{locnarr} for pre-training belong to the \nocapsxd{} leaderboard rather than the default one. 

Some of our results on the \cc{} benchmark are taken from the leaderboard, which is located at \url{https://ai.google.com/research/ConceptualCaptions/leaderboard?active_tab=leaderboard}.

\subsection{Hyperparameter search}

For pre-training experiments, we do not conduct hyperparameter tuning besides an initial stage of exploration as we believe small changes would not considerably affect the downstream performance. For instance, we fix an initial learning rate to 0.000032 and observe it works consistently well (on the validation set) across scenarios.

For fine-tuning experiments, we focus on tuning one hyperparamter: the initial learning rate. In the case of \nocaps{}, we also lightly tune the maximum number of training steps as we observe the model overfitting on \cococap{}. In all cases, we make sure to allocate similar resources to any two settings that we make a comparison between, such as pre-training data sources of \cc{} and \ccp{}.

For generation, the ranges for the initial learning rate are \{3.2e-9, 3.2e-8, 3.2e-7\} and the ranges for the maximum number of training steps are \{5K, 10K\}.
For matching, the ranges for the initial learning rate are \{3.2e-8, 3.2e-7, 3.2e-6\} while the maximum number of training steps is fixed to 10K.

\end{document}